\PassOptionsToPackage{table}{xcolor}
\documentclass{article} %
\usepackage[numbers,sort&compress]{natbib}
\usepackage{iclr2027_conference,times}

\usepackage{amsmath,amsfonts,bm}

\def\eqref#1{equation~\ref{#1}}
\def\1{\bm{1}}

\DeclareMathAlphabet{\mathsfit}{\encodingdefault}{\sfdefault}{m}{sl}
\SetMathAlphabet{\mathsfit}{bold}{\encodingdefault}{\sfdefault}{bx}{n}

\usepackage{hyperref}
\usepackage{url}
\usepackage{amsmath}
\usepackage{float,psfrag,epsfig,color,xcolor,url}
\usepackage{amsmath, amssymb}
\usepackage[utf8]{inputenc} %
\usepackage[T1]{fontenc}    %
\usepackage{multirow}
\usepackage{xcolor}
\usepackage{url}            %
\usepackage{booktabs}       %
\usepackage{amsfonts}       %
\usepackage{nicefrac}       %
\usepackage{microtype}      %
\usepackage{xcolor}         %
\usepackage{subcaption}
\definecolor{reasoningblue}{HTML}{2E5C8A}
\definecolor{answerorange}{HTML}{D97706}
\definecolor{decorationgreen}{HTML}{059669}
\usepackage{pifont}
\usepackage{mathtools}
\usepackage{amsthm}
\usepackage{newunicodechar}
\usepackage{graphicx}
\graphicspath{{./}}
\usepackage{wrapfig}
\usepackage[most]{tcolorbox}
\definecolor{takeawayblue}{RGB}{85,135,195}
\definecolor{takeawaybg}{RGB}{247,248,255}
\usepackage[table]{xcolor}
\newtcolorbox{takeawaybox}{
    enhanced,
    breakable,
    colback=takeawaybg,
    colframe=takeawayblue,
    colbacktitle=takeawayblue,
    coltitle=white,
    title=Takeaway,
    fonttitle=\bfseries\normalsize,
    boxrule=1pt,
    arc=3mm,
    left=8pt,
    right=8pt,
    top=8pt,
    bottom=8pt,
    toptitle=1pt,
    bottomtitle=2pt,
    before skip=10pt,
    after skip=10pt
}

\newtheorem{proposition}{Proposition}

\newtheorem{theorem}{Theorem}
\newtheorem{corollary}{Corollary}

\newcommand{\uc}[1]{{\scriptsize\texttt{[U+#1]}}}
\DeclareUnicodeCharacter{FFFD}{\ensuremath{\square}}  %
\DeclareUnicodeCharacter{FEFF}{}                      %
\DeclareUnicodeCharacter{05E7}{\uc{05E7}}
\DeclareUnicodeCharacter{2744}{\uc{2744}}
\DeclareUnicodeCharacter{2705}{\uc{2705}}
\DeclareUnicodeCharacter{267B}{\uc{267B}}
\DeclareUnicodeCharacter{2699}{\uc{2699}}
\tcbuselibrary{breakable}
\usepackage{fvextra}

\usepackage{etoolbox}
\usepackage{etoolbox}
\makeatletter
\patchcmd{\@maketitle}{\lhead{Published as a conference paper at ICLR 2027}}{}{}{}
\patchcmd{\@maketitle}{\lhead{Under review as a conference paper at ICLR 2027}}{}{}{}
\makeatother
\renewcommand{\headrulewidth}{0pt}

\usepackage{tikz}
\hypersetup{
  hidelinks,
  pdftitle={When Students Fail to Learn from Teachers: Learning Dynamics of On-Policy Distillation},
  pdfauthor={Xinnan Zhang, Zeyi Liao, Siliang Zeng, Zhongruo Wang, Jiaxiang Li, Xinyan Zhao, Huan Sun, Mingyi Hong},
  pdfsubject={},pdfkeywords={}
}
\makeatletter
\patchcmd{\@maketitle}{\vskip 0.3in minus 0.1in}{\vskip 8pt}{}%
  {\PackageError{preprint}{Title spacing could not be applied}{Check the title macro.}}
\makeatother

\newcommand{\projectwebsiteurl}{}  %
\newcommand{\projectrepository}{https://github.com/zhangxinnan123/learning-dynamic-opd}
\newcommand{\projectrepositoryname}{learning-dynamic-opd}
\definecolor{projectlink}{HTML}{315C83}
\newcommand{\projectglobe}{\tikz[baseline=-.55ex,scale=.13,line width=.6pt,color=black]{%
  \draw (0,0) circle (1);\draw (0,0) ellipse (.42 and 1);
  \draw (-1,0)--(1,0);\draw (-.87,.5)--(.87,.5);\draw (-.87,-.5)--(.87,-.5);}}
\newcommand{\projectlinks}{%
  \ifboolexpr{test {\ifdefempty{\projectwebsiteurl}} and test {\ifdefempty{\projectrepository}}}{}{%
    \noindent\makebox[\textwidth][c]{\small\color{projectlink}%
      \ifdefempty{\projectwebsiteurl}{}{\href{\projectwebsiteurl}{\projectglobe\, Website}}%
      \ifboolexpr{test {\ifdefempty{\projectwebsiteurl}} or test {\ifdefempty{\projectrepository}}}{}%
        {\hspace{1em}\textperiodcentered\hspace{1em}}%
      \ifdefempty{\projectrepository}{}{\href{\projectrepository}{\texttt{\projectrepositoryname}}}}\par
    \vspace{3pt}}}

\title{When Students Fail to Learn from Teachers:\\Learning Dynamics of On-Policy Distillation}

\author{%
Xinnan Zhang\textsuperscript{1*}\quad
Zeyi Liao\textsuperscript{2*}\quad
Siliang Zeng\quad
Zhongruo Wang\quad
Jiaxiang Li\textsuperscript{4}\\
\bfseries Xinyan Zhao\textsuperscript{3}\quad
Huan Sun\textsuperscript{2}\quad
Mingyi Hong\textsuperscript{1}\\[4pt]
\normalfont\textsuperscript{1}University of Minnesota\quad
\textsuperscript{2}The Ohio State University\quad
\textsuperscript{3}Amazon\quad
\textsuperscript{4}Virginia Tech\\
\normalfont\texttt{zhan9359@umn.edu}
}

\iclrfinalcopy
\begin{document}

\maketitle
\begingroup\renewcommand\thefootnote{}\footnotetext{\textsuperscript{*}Equal contribution.}\endgroup
\fancyhead{}
\renewcommand{\headrulewidth}{0pt}

\projectlinks

\begin{abstract}
On-policy distillation (OPD) has become a core technique for post-training large language models, enabling students to learn from dense token-level supervision on their own generations. However, even with a capable teacher, OPD training can be unstable, leading to excessive response length, repetitive generation, and severe performance degradation. One example is \textit{spurious agreement}, where the student maintains high overlap with the teacher at the states visited but produces repetitive generations. Although such failures have been observed, the mechanisms behind these failures remain poorly understood. To address this gap, we investigate OPD through the lens of learning dynamics, tracing three mechanisms from local analysis to subsequent on-policy sampling. We first identify states where sampled-token OPD updates fail to correct teacher--student mismatch and characterize them using the student's mass on the teacher's top-$p$ set. We then show that learning from the teacher is difficult at mismatched states and that their gradients can also affect with learning at other tokens. Finally, using a termination-token mismatch as a case study, we study how continued training on mismatched states influences on-policy sampling, suppressing the student's current modes and leading the student toward states where the teacher is unreliable. These findings help explain the benefits of supervised fine-tuning (SFT) initialization and motivate targeted correction of states where student and teacher are mismatched. We introduce state-dependent KL routing, which applies forward KL when student coverage of teacher-preferred tokens is low or a substantial fraction of teacher probability mass is underestimated. Across three Qwen3 model scales and both Base and SFT initialization, our method improves mean Avg@8 by 1.38--2.50 percentage points and mean Best@8 by 1.01--6.24 points over sampled-token OPD on mathematical reasoning benchmarks, and improves mean Avg@8 by 2.92--3.44 points and mean Best@8 by 1.27--2.88 points on code generation benchmarks.

\end{abstract}

\section{Introduction}
On-policy distillation (OPD)~\citep{agarwal2024policy} has become an important part of large language model (LLM) post-training, which is widely used in Qwen3~\citep{yang2025qwen3}, Mimo~\citep{xiao2026mimo}, DeepSeek-V4~ \citep{xu2026deepseek}. In OPD, a student generates its own responses and learns from a teacher's token-level feedback conditioned on the resulting prefixes~\citep{ross2011reduction}. Unlike group-based reinforcement learning (RL) algorithms, such as GRPO~\citep{shao2024deepseekmath}, OPD provides token-level dense reward, making it compute-efficient and effective~\citep{lu2025onpolicydistillation}. 

\begin{figure}[h]
    \centering
    \includegraphics[width=1.0\linewidth]{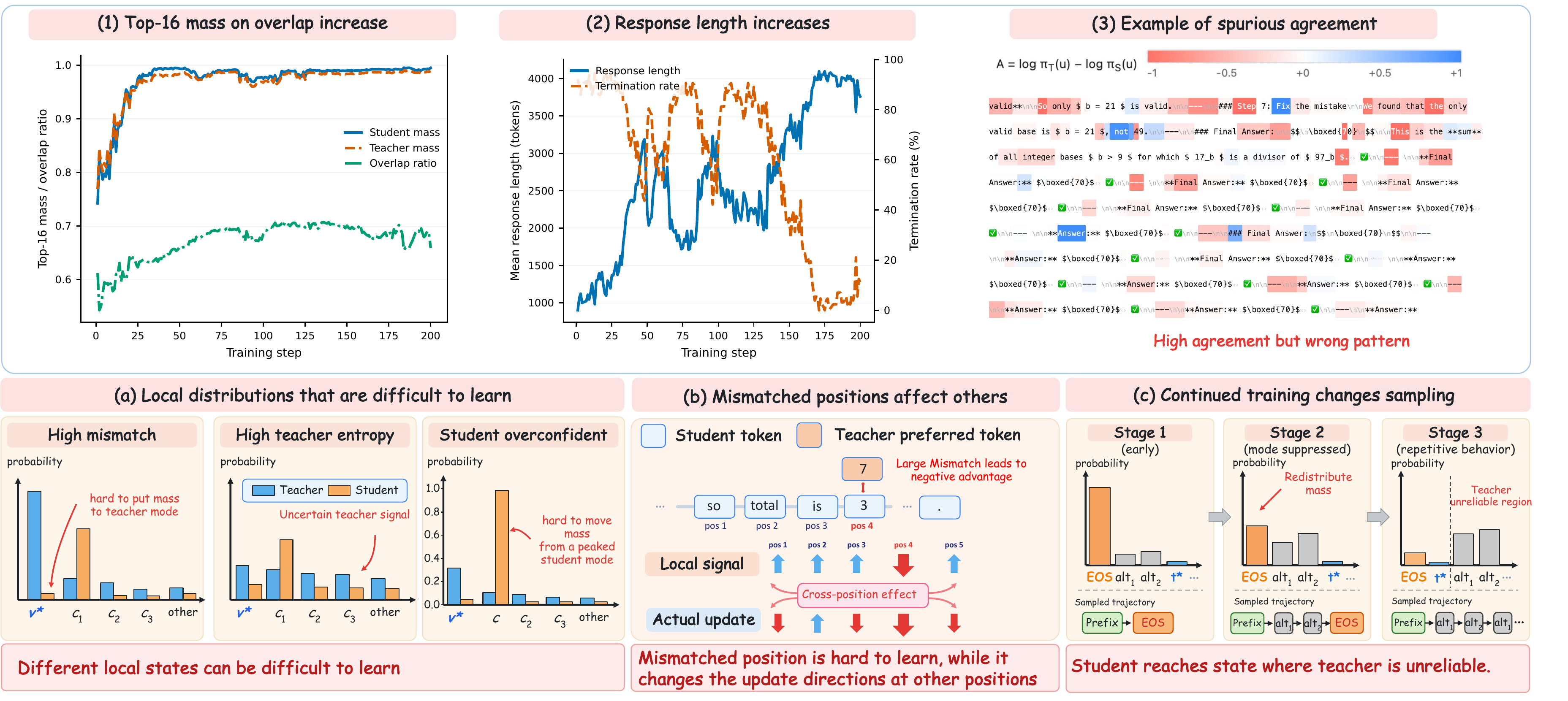}
    \caption{Learning dynamics of OPD. Overlap definitions are given in Appendix~\ref{app:diagnostic-setup}.}
    \label{fig:motivation}
\end{figure}

However, even with a strong teacher, OPD does not always lead to effective student learning and can suffer from instability problems, such as repetitive generation and severe performance degradation~\citep{fu2026revisiting}. For example, 
as shown in Figure~\ref{fig:motivation}, OPD training leads to the phenomenon that we term \emph{spurious agreement}: the student maintains high overlap mass with the teacher as training progresses but its generation is repetitive, leading to excessive response length.

Prior work has found these failures and proposed ways to improve the stability of OPD, such as incorporating forward KL~\citep{fu2026revisiting} or adding a supervised fine-tuning (SFT) objective~\citep{luo2026demystifying}. In addition, other studies examine which states are useful for OPD training~\citep{armandpour2026unmasking,zhu2026many}. These include selecting informative tokens~\citep{xu2026tip,wang2026not}, restricting updates within the teacher's trust region~\citep{xing2026trust}, and selectively applying forward KL based on conditional information such as teacher entropy~\citep{jin2026entropy} or advantage~\citep{jia2026asymmetric}. While these studies improve the stability of OPD, they do not provide a principled explanation of how token-level teacher feedback contributes to the policy updates and how these updates affect subsequent learning. 

\textbf{Research Question, Proposed Work and Contributions.} Based on the above discussion, this work focuses on addressing the following key research question:
	\begin{center}
		\vspace{-0.5cm}
		\setlength\fboxrule{0.0pt}
		\noindent\fcolorbox{black}[rgb]{0.95,0.95,0.95}{\begin{minipage}{0.99\columnwidth}
        \begin{center}
            	    {Given a strong teacher, why does a student sometimes fail to learn from the teacher's feedback during OPD, and how can we improve learning?}
        \end{center}
		\end{minipage}}
		\vspace{-0.2cm}
	\end{center}

To answer this question, we study OPD through the lens of learning dynamics~\citep{ren2025learning}, from a softmax policy to practical LLM training. We first examine how sampled-token updates change the student's token distribution at the given state and identify states where teacher--student mismatch remains difficult to correct.
We then study how gradients from different token positions contribute to the final distribution change, focusing on how different positions affect learning at other positions. Finally, we examine how continued training changes subsequent on-policy sampling, connecting local policy updates to the generation failures illustrated in Figure~\ref{fig:motivation}.
Our main contributions are summarized as follows:

\noindent \textbf{(1)}
To begin with, we study how sampled-token updates change the student's distribution at a given state. Our theoretical analysis under a single-state softmax policy characterizes the direction and magnitude of token probability changes, showing why a student may fail to learn from the teacher or even worsen after updating.

\noindent {\bf (2)}
We investigate how these learning dynamics affect practical LLM training. We identify states where sampled-token updates provide insufficient supervision and show how the student's probability mass on the teacher's top-$p$ set helps characterize them. We further show that states with high mismatch can be difficult for the student to learn from while their gradients will affect learning at other positions. Finally, we examine how teacher--student mismatch affects subsequent on-policy sampling. Using EOS-token mismatch as a case, we show that continued training leads to spurious agreement.

\noindent {\bf (3) }
Based on the above findings, we show that SFT provides a favorable initialization for OPD and introduce state-dependent KL routing to address low student coverage of teacher-preferred tokens or underestimation of a substantial fraction of teacher probability mass. Across Qwen3 model scales and mathematical reasoning benchmarks, our approach improves training stability and performance over sampled-token OPD and the baselines.

\section{Preliminaries}

\paragraph{On-policy Distillation.}
The objective of on-policy distillation (OPD) is to let a student model imitate the teacher model's behavior
using on-policy data. Specifically, given a prompt $x \sim \mathcal{D}_x$, the trajectory $y = (y_1, y_2, ...,y_T)$ is sampled from the student model with a total of $T$ generated tokens. The objective of OPD is to minimize the sequence-level reverse KL over student-generated trajectories. By the autoregressive factorization, this objective decomposes as
\begin{align}
\mathcal L_{\mathrm{OPD}}(\theta)
&=
\mathbb E_{\substack{x\sim\mathcal D_x,
y\sim\pi_\theta(\cdot\mid x)}}
\left[
\sum_{t=1}^{T}
D_{\mathrm{KL}}\left(
\pi_\theta(\cdot\mid s_t)
\|\pi_T(\cdot\mid s_t)
\right)
\right],
\label{eq:prelim}
\end{align}
where $s_t=(x,y_{<t})$ is a student-visited state.

In practice, implementations vary in how the reverse KL is computed. The methods can be categorized into three classes: (1) full-vocabulary OPD computes the divergence over the full vocabulary; (2) partial OPD restricts the sum to a subset, such as the teacher's or student's top-$K$ tokens; and (3) sampled-token OPD uses student-sampled tokens to obtain a single-sample Monte Carlo estimator. The latter avoids an explicit vocabulary-wide KL computation and is the focus of this work. 

Following \citet{lu2025onpolicydistillation}, the resulting sampled-token OPD gradient is
\begin{equation}
\nabla_\theta \mathcal{L}_{\mathrm{opd}}(\theta)
=-\mathbb{E}_{y \sim \pi_\theta}\!\left[\sum_{t=1}^{T} A_t\,\nabla_\theta \log \pi_\theta(y_t\mid s_t)\right],
\label{eq:prelim:opd-grad}
\end{equation}
where $A_t=\log\pi_T(y_t\mid s_t)-\log\pi_\theta(y_t\mid s_t)$ denotes the token-level advantage. Note that Eq.~\ref{eq:prelim:opd-grad} is not the exact gradient of Eq.~\ref{eq:prelim}. It keeps only the immediate term $A_t$ of the reward-to-go, which is equivalent in expectation to treating the distribution of student-visited prefixes as fixed (Appendix~\ref{app:objective-relation}) Eq.~\ref{eq:prelim} is thus the sequence-level objective that motivates OPD, while Eq.~\ref{eq:prelim:opd-grad} is the update used and analyzed in this work.

\section{Policy Dynamics Analysis}
\label{sec:dynamics}

In this section, we first present a theoretical and empirical analysis of how OPD updates change token probabilities under a single-state softmax policy. 

At a fixed state $s$, the student policy with the softmax parametrization can be written as
\begin{equation}
\label{eq:softmax}
\pi_\theta^k(a\mid s)
=
\frac{\exp(\theta_{s,a}^k)}
{\sum_{b\in\mathcal V}\exp(\theta_{s,b}^k)},
\end{equation}
where $\mathcal{V}$ is the vocabulary set, and $k$ is the iteration index. Let $A(s,u) := \log\pi_T(u|s) - \log\pi_\theta^k(u|s)$ denote the advantage, a single-sample SGD step on the sampled-token OPD loss updates the logits as
\begin{equation}
\label{eq:opd-update}
\theta_s^{k+1}
\;=\;
\theta_s^k \;+\; \eta\,A(s,u)\,
\bigl(\mathbf e_u - \pi_\theta^k(\cdot|s)\bigr),
\qquad u\sim\pi_\theta^k(\cdot|s),
\end{equation}
where $\mathbf e_u \in \mathbb{R}^{|\mathcal V|}$ is the one-hot vector for the sampled token $u$.
 
We characterize the resulting one-step dynamics through the differences:
\begin{equation}
\label{eq:delta_pi}
\Delta\log\pi(c\mid s)
:= \log\pi_\theta^{k+1}(c\mid s)-\log\pi_\theta^k(c\mid s),
\qquad
\Delta\pi(c\mid s)
:= \pi_\theta^{k+1}(c\mid s)-\pi_\theta^k(c\mid s).
\end{equation}

\begin{proposition}[Per-token training dynamics under OPD]
\label{prop:per-token}
After one update~\eqref{eq:opd-update}, for every $c\in\mathcal V$,
\begin{equation}
\label{eq:dpi-master}
\Delta\log\pi(c|s) = \eta\,A(s,u)\,B(c,u;s) + O(\eta^2),
\quad
\Delta\pi(c|s) = \pi_\theta^k(c|s)\,\Delta\log\pi(c|s) + O(\eta^2),
\end{equation}
where
\begin{equation}
\label{eq:bracket-def}
B(c,u;s) \;:=\;
\sum_{b\in\mathcal V}\bigl(\mathbf 1\{c=b\}-\pi_\theta^k(b|s)\bigr)
\bigl(\mathbf 1\{u=b\}-\pi_\theta^k(b|s)\bigr).
\end{equation}
\end{proposition}

The proof is presented in Appendix~\ref{app:proof-prop1}. The factorization decomposes the policy dynamics into three parts: the advantage $A(s,u)$, the probability weight $\pi_\theta^k(c|s)$, and the coupling term
$B(c,u;s)$ that captures the local distribution. We next analyze Prop.~\ref{prop:per-token} in terms of update direction and magnitude.

\paragraph{Direction of token probability changes.}
\label{sec:direction}

Let
$p:=\pi_\theta^k(u\mid s)$ and
$S_{-u}(s):=\sum_{b\neq u}\pi_\theta^k(b\mid s)^2$.
For the sampled token $c=u$,
\begin{equation}
\label{eq:Buu}
B(u,u;s)
=
1-2p+\|\pi_\theta^k(\cdot\mid s)\|_2^2
=
(1-p)^2+S_{-u}(s)
\ge 0.
\end{equation}
Therefore, with a  sufficiently small learning rate $\eta$, the sampled token always moves in the direction of advantage, i.e., $\operatorname{sign}
\bigl(
\Delta\pi(u\mid s)
\bigr)
=
\operatorname{sign}
\bigl(
A(s,u)
\bigr)$.
The update also redistributes probability mass among non-sampled tokens through the softmax normalization. For the token $c\neq u$,
\begin{equation}
\label{eq:bracket-noneq}
B(c,u;s)
=
\tau(u;s)-\pi_\theta^k(c\mid s),
\end{equation}
where $\tau(u;s)
:=
\|\pi_\theta^k(\cdot\mid s)\|_2^2
-
\pi_\theta^k(u\mid s)$.
Hence, non-sampled tokens on opposite sides of $\tau(u;s)$ move in opposite directions. Unlike the token $u$, their movement is
therefore determined not only by the advantage but also by their position in the current student distribution. 

\paragraph{Magnitude of token probability changes.}
\label{sec:amplitude}

We next analyze the magnitude of token probability change. From~\eqref{eq:Buu}, we have
\begin{equation}
\label{eq:dpi-u-clean}
\Delta\pi(u\mid s)
=
\eta A(s,u)\,
p\bigl[(1-p)^2+S_{-u}(s)\bigr]
+
O(\eta^2).
\end{equation}
Since $0
\le
S_{-u}(s)
\le
(1-p)^2$,
the first-order update magnitude satisfies
\begin{equation}
\label{eq:dpi-u-bounds}
|A(s,u)|p(1-p)^2
\;\le\;
\lim_{\eta\to0}
|\Delta\pi(u\mid s)|/\eta
\;\le\;
2|A(s,u)|p(1-p)^2.
\end{equation}

The sampled-token probability change is therefore jointly controlled by the advantage and the token's probability. In particular, even when $|A(s,u)|$ is large, the effective probability change becomes small when the sampled token lies near either boundary: the factor $p$ limits the magnitude of probability changes for low-probability tokens, while $(1-p)^2$ limits them for high-probability tokens.

For a non-sampled token $c\neq u$, the change of probability can be written as
\begin{equation}
\label{eq:dpi-c}
|\Delta\pi(c\mid s)|
=
\eta |A(s,u)|\,
\pi_\theta^k(c\mid s)
\left|
\tau(u;s)-\pi_\theta^k(c\mid s)
\right| + O(\eta^2),
\end{equation}
which depends on the advantage, token's
probability and $\tau(u;s)$. In particular, low-probability tokens receive little probability change. We defer a more detailed analysis to Appendix~\ref{app:softmax-details}.

\paragraph{Takeaway.}
These analyses explain why a student may fail to learn from a strong teacher. First, tokens preferred by the teacher may change little when their student probabilities are low. Second, non-sampled tokens may change in directions that do not match the teacher's preferences. Third, the sampled-token advantage compares the teacher's and student's probabilities for only one token. It can be small even when their full distributions differ greatly. The point also holds for full-vocabulary reverse OPD~\citep{xu2026deepseek}, whose update is the expectation of the sampled-token update and is still scaled by the student probability of each token (Appendix~\ref{app:full-vocab}).

\section{Diagnosing OPD from learning dynamics}
Building on the above analysis, we investigate why a student may fail to learn from a strong teacher in practical LLM training. First, we study local distribution dynamics to understand when sampled-token OPD updates make it hard for the student to learn from the teacher. We then examine cross-position effects, focusing on how gradients from different token positions affect actual update directions. Finally, we study on-policy sampling dynamics to understand how
continued training changes the student's generation behavior.
Experimental settings and details are provided in Appendix~\ref{app:diagnostic-setup}.

\subsection{Local Distributional Dynamics}

\paragraph{Token probability changes.}

\begin{figure}[t]
    \centering
    \captionsetup{skip=5pt}
    \captionsetup[subfigure]{skip=3pt}
    \begin{subfigure}[b]{0.32\linewidth}
        \centering
        \includegraphics[width=\linewidth]{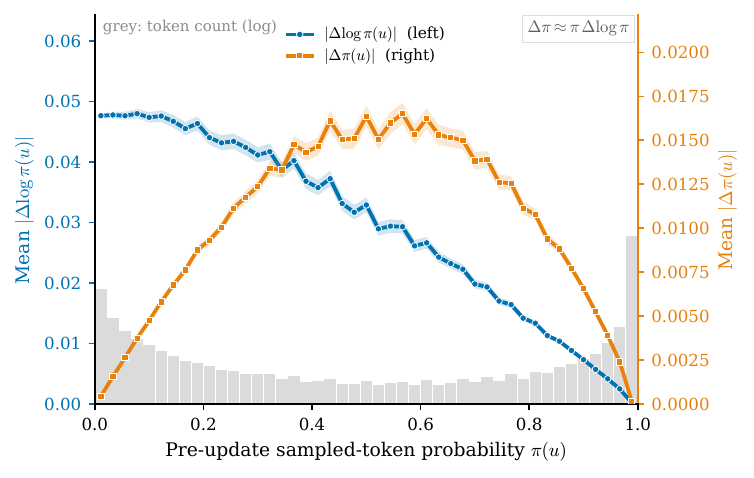}
        \caption{}
        \label{fig:l2-mobility}
    \end{subfigure}
    \hfill
    \begin{subfigure}[b]{0.32\linewidth}
        \centering
        \includegraphics[width=\linewidth]{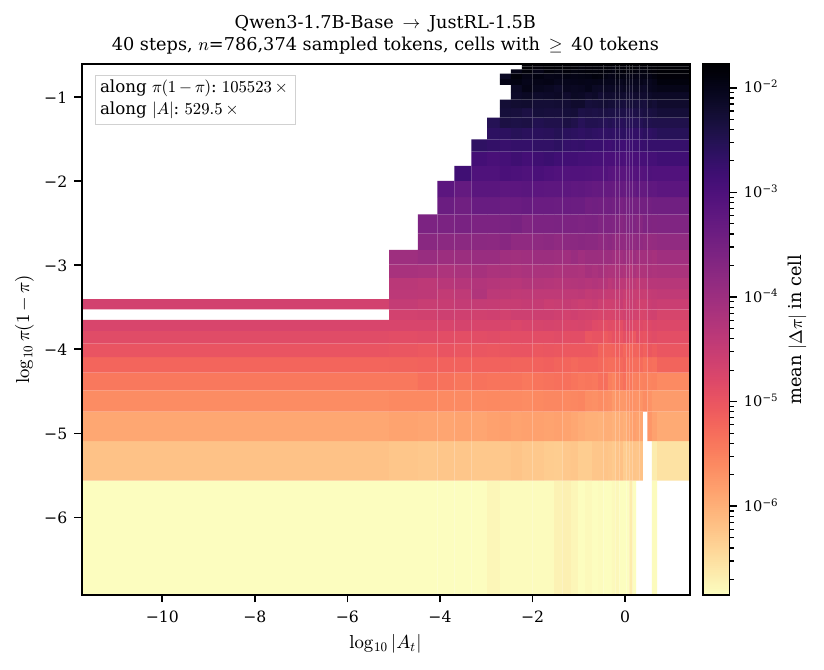}
        \caption{}
        \label{fig:l2-predict}
    \end{subfigure}
        \hfill
    \begin{subfigure}[b]{0.32\linewidth}
        \centering
        \includegraphics[width=\linewidth]{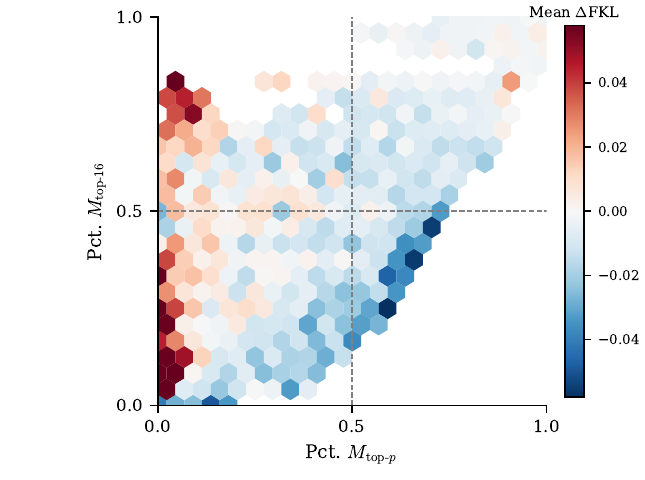}
        \caption{}
        \label{fig:l4-predict}
    \end{subfigure}
\caption{
\textbf{Local distributional dynamics under OPD.}
(a) Log-probability and probability-mass changes after one training step grouped by student probability.
(b) Mean $|\Delta\pi|$ within joint bins of
$|A|$ and $\pi(1-\pi)$.
(c) Mean $\Delta\mathrm{FKL}$ jointly grouped by
student coverage of the teacher's top-$p$ ($p=0.9$) and top-$K$ (K=16) sets. Blue indicates improved alignment; red indicates deterioration.
}
\label{fig:level2}

\end{figure}

The above analysis shows that the token's current probability
affects the magnitude of its probability change. We first examine whether this pattern also appears in practical LLM training. As shown in Figure~\ref{fig:level2}(a), low-probability sampled tokens can have large log-probability changes but small probability changes, whereas $|\Delta\pi|$ peaks at intermediate student probabilities.  Figure~\ref{fig:level2}(b) further shows that probability change varies
with $\pi(1-\pi)$ even within similar ranges of $|A|$. Appendix~\ref{app:token-movement} extends this analysis to non-sampled tokens and compares probability- and advantage-based predictors. These results indicate that token probability changes are small when the student's current probability is either low or high. In particular, when the student assigns little probability mass to teacher-preferred
tokens, sampled-token updates fail to learn the teacher's preferred token.

\paragraph{Top-$p$ coverage characterizes distributional mismatch.}

We next examine whether these cumulative changes bring the student distribution closer to the teacher's. We measure teacher--student mismatch using forward KL (FKL). For a prefix $s_{i}=(x,y_{<i})$, we define
\begin{equation}
\mathrm{FKL}_i
 = \sum_{v \in \mathcal{V}}\pi_T(v\mid s_{i})(\log \pi_T(v\mid s_{i}) - \log \pi_\theta(v\mid s_{i})).
\end{equation}
To isolate changes in the student's distribution from changes in the states it visits, we use a held-out set of fixed teacher-generated prefixes. We evaluate the student on the same prefixes before training and after multiple OPD updates, and report the change
$\Delta\mathrm{FKL}_i
=\mathrm{FKL}_i^{\mathrm{after}}-\mathrm{FKL}_i^{\mathrm{initial}}$.
Negative values indicate improved alignment, while positive
values indicate deterioration. 

To better understand teacher--student mismatch, we measure how much student probability lies on the teacher’s high-probability tokens. Let $\mathcal S_p(s)$ be the smallest set of highest-probability teacher tokens whose cumulative
teacher mass reaches $p$, and let $\mathcal S_K(s)$
contain the teacher's $K$ highest-probability tokens. We define the student's coverage of these sets as
\begin{equation}
\label{eq:mass_support}
M_{\mathrm{top}\text{-}p}=
\sum_{c\in\mathcal{S}_p(s)}
\pi_\theta(c\mid s), \quad M_{\mathrm{top}\text{-}K}=
\sum_{c\in\mathcal{S}_K(s)}
\pi_\theta(c\mid s).
\end{equation} 
Low coverage indicates that the student assigns little probability mass to teacher-preferred tokens. Unlike top-$K$,
top-$p$ adjusts the number of selected tokens based on the
teacher's distribution.

Figure~\ref{fig:level2}(c) shows that deterioration concentrates at low top-$p$ coverage across a broad range of top-$K$ coverage. 
This suggests that top-$p$ coverage is more informative for capturing the mismatch between student and teacher. We therefore use top-$p$ coverage in the following analyses.

\paragraph{Characterizing states that are difficult to learn from.}

We next study which states are difficult for the student to learn from. We measure the change in FKL at fixed
held-out states before and after OPD and analyze states
in the highest and lowest initial FKL quartiles separately. We characterize these states using student and teacher entropy, and the student's coverage of the teacher's top-$p$ set.

\begin{figure}[h]
    \centering
    \includegraphics[width=1.0\linewidth]{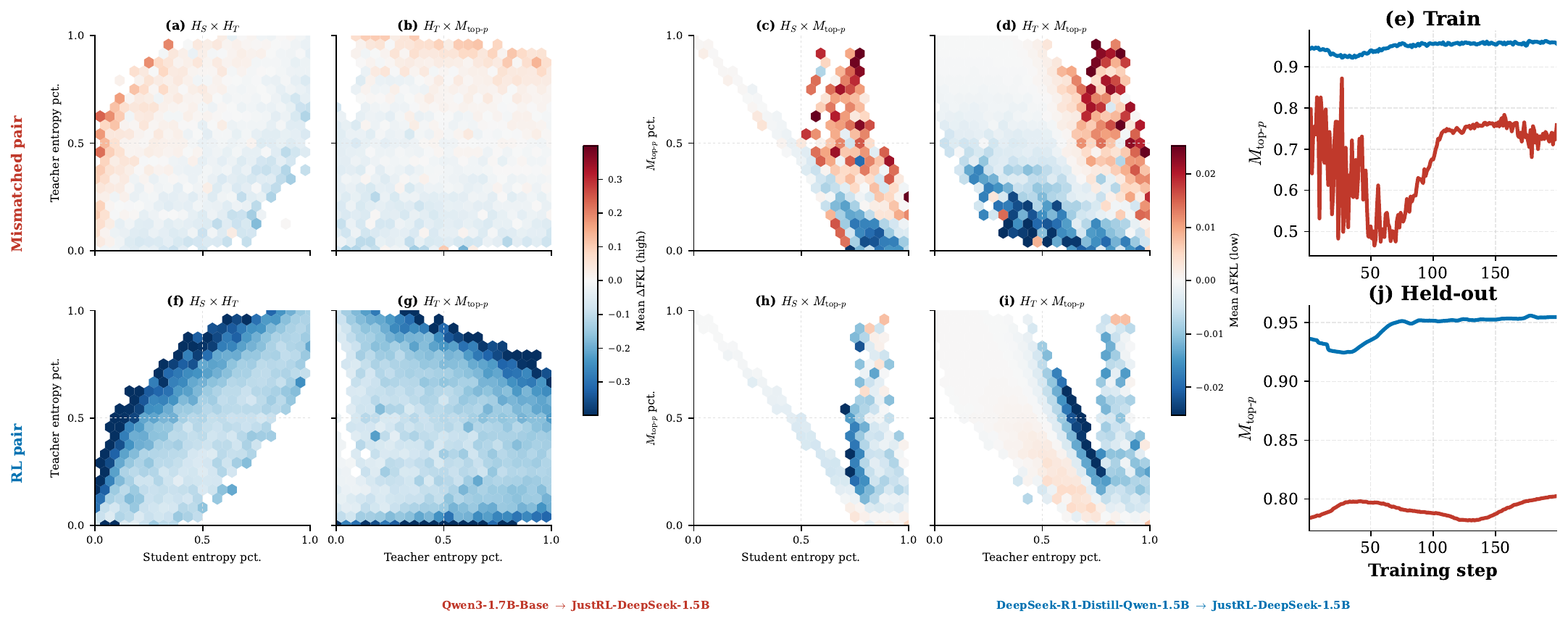}
    \vspace{-0.5cm}
    \caption{
    \textbf{Local learning dynamics under two teacher--student pairs.} Rows show the mismatched and RL-aligned pairs. Columns 1--2 and 3--4 show changes in FKL at fixed held-out states with high and low initial FKL, respectively. The final column tracks $M_{\mathrm{top}\text{-}p}$ during training.
    }
    \label{fig:local_regime}
\end{figure}

We first compare two students with the same teacher, \texttt{JustRL-1.5B}, an RL-trained version of \texttt{DeepSeek-R1-Distill-Qwen-1.5B} (R1-Distill-1.5B). As shown in Figure~\ref{fig:local_regime}(e,j), the R1-Distill-1.5B student maintains high coverage on both training and held-out states, while the Qwen3-1.7B-Base student has substantially lower coverage. As a result, for the Qwen student, FKL increases at states with
high mismatch and low student entropy
(Figure~\ref{fig:local_regime}(a)). At states with low
top-$p$ coverage, even when FKL decreases, the improvement is small relative to the large initial mismatch distance (Figure~\ref{fig:local_regime}(b)). When initial mismatch is low, changes are generally smaller, and the student becomes worse in regions with high teacher entropy. In contrast, the R1-Distill-1.5B student learns well at different states.

\begin{wrapfigure}{r}{0.48\columnwidth}
    \centering
    \vspace{-8pt}
    \includegraphics[width=\linewidth]{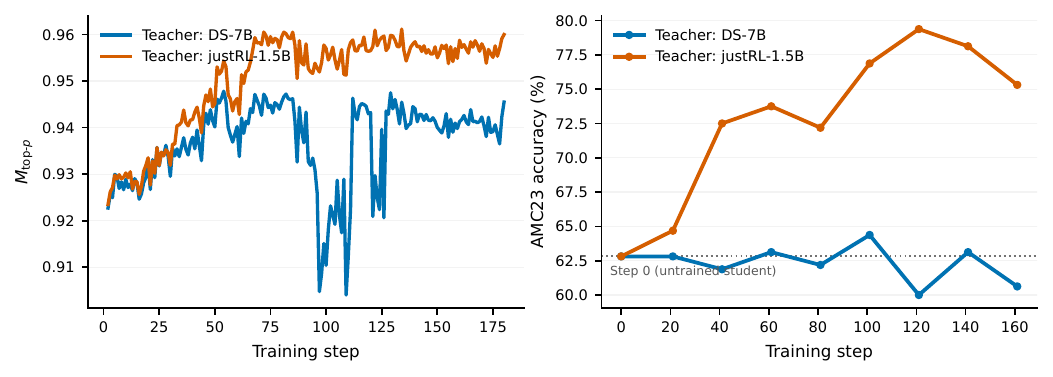}
    \caption{
    Same student (R1-Distill-1.5B) with different teachers, 
DeepSeek-Distill-Qwen-7B (DS-7B) and JustRL-1.5B.
    }
    \label{fig:teacher-comparison}
    \vspace{-8pt}
\end{wrapfigure}

As a complementary comparison, we fix the student and vary the teacher (Fig.~\ref{fig:teacher-comparison}). When the student assigns more probability mass to the teacher's top-$p$ set, this coverage further increases during training and AMC23 accuracy improves substantially. With lower coverage, the student shows little improvement. This further supports the importance of student coverage for learning from teacher feedback.

These results demonstrate the importance of student coverage of the teacher's distribution for learning under sampled-token OPD. When student coverage is high, it generally learns well from the teacher's supervision. When student and teacher have substantial mismatch, the gap can increase at states with high teacher entropy and low student entropy. At low-coverage states, FKL may
decrease, but the improvement is small relative to the large initial mismatch. We next examine how cross-position contributions can further affect the updates induced by teacher feedback.

\subsection{Cross-Position Effects}

The softmax analysis treats logits at each state
as independent parameters; however, this assumption no longer holds in an LLM. In an LLM, logits across positions are produced by shared parameters, so an update at one position also affects others. We study how these effects change the student's actual updates during OPD.

\paragraph{Token update direction.}

\begin{figure}[h]
    \centering
    \captionsetup{skip=4pt}
    \begin{minipage}[c]{0.22\textwidth}
        \centering
        \includegraphics[width=\linewidth]{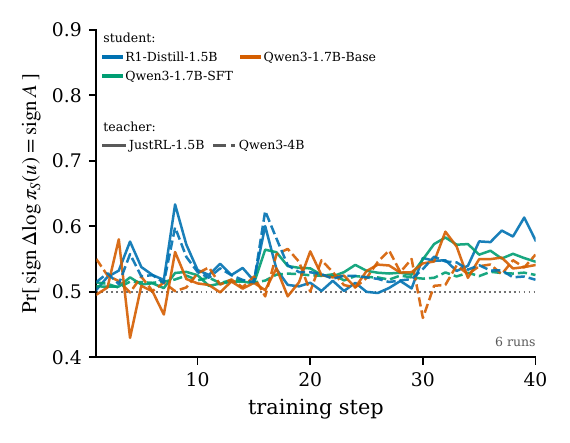}
    \end{minipage}\hfill
    \begin{minipage}[c]{0.22\textwidth}
        \centering
        \includegraphics[width=\linewidth]{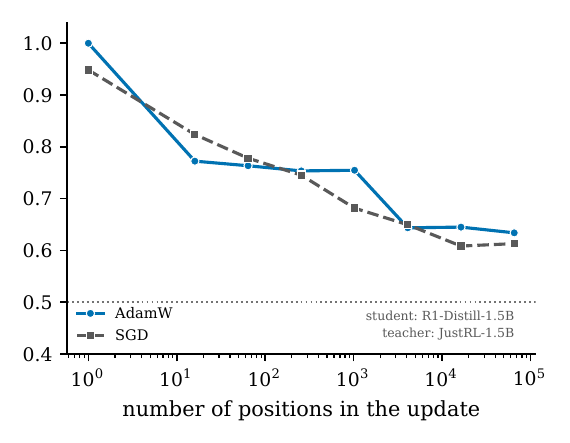}
    \end{minipage}\hfill
    \begin{minipage}[c]{0.22\textwidth}
        \centering
        \includegraphics[width=\linewidth]{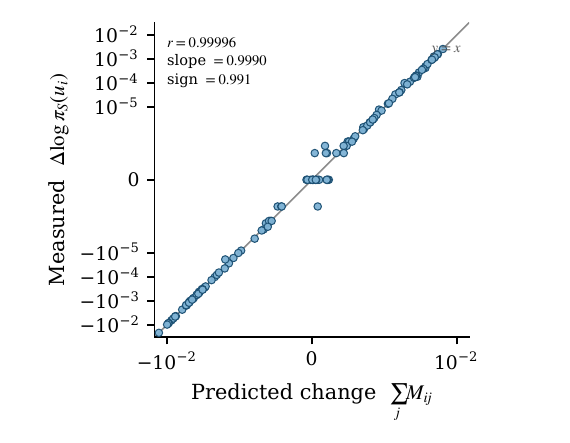}
    \end{minipage}\hfill
    \begin{minipage}[c]{0.27\textwidth}
    \centering
    \resizebox{\linewidth}{!}{%
        \begin{tabular}{@{}clrr@{}}
            \toprule
            Rank & Token & $A$ & Weight \\
            \midrule
            1 & \texttt{<|endoftext|>} & $-22.53$ & 31.81\% \\
            2 & \texttt{Rightarrow} & $3.76$ & 10.09\% \\
            3 & \texttt{that} & $2.09$ & 6.20\% \\
            4 & \texttt{become} & $-2.61$ & 3.55\% \\
            5 & \texttt{terms} & $-1.35$ & 3.03\% \\
            \bottomrule
        \end{tabular}%
    }
\end{minipage}
    \caption{
\textbf{Cross-position effects on update directions.}
(a) Sign agreement between $A_i$ and
$\Delta\log\pi_\theta(u_i\mid s_i)$ across training steps. (b) Sign agreement as the number of positions included in an update increases. (c) First-order predictions $\sum_jM_{ij}$ versus measured changes after one SGD step. (d) Top 5 tokens by absolute cross-position influence $\sum_{i\neq j}|M_{ij}|$. Weights are these influences normalized to sum to $100\%$ over the selected 200 positions.}
    \label{fig:batch}
    \vspace{-8pt}
\end{figure}

In Prop.~\ref{prop:per-token}, for a sufficiently
small update, the sampled token changes in the direction of its advantage, e.g., $\operatorname{sign}\!\left(\Delta\log\pi_S(u_i)\right)
=\operatorname{sign}(A_i)$. However, in LLM training, the two signs agree for only around $50\%$ across different teacher--student pairs (Figure~\ref{fig:batch}(a)).
To understand this difference, we vary the number of token positions included in an update.
The sign agreement between advantage and change is close to $100\%$ when only one position is included, but decreases as more positions are added
(Figure~\ref{fig:batch}(b)).
This indicates that the gradients from other positions can change the update direction (see Appendix~\ref{app:interaction-validation} for additional analysis).

For a batch of tokens $\mathcal B$,
let $h_i=\nabla_\theta\log\pi_\theta(u_i\mid s_i)$
denote the log-probability gradient. A single SGD update gives
$\Delta\theta=\eta\sum_{j\in\mathcal B}A_jh_j$.
We define the first-order effect of position $j$ on position $i$ as $M_{ij}=\eta A_j\langle h_i,h_j\rangle$. The resulting change at position $i$ is
\small
\begin{equation}
\label{eq:cross-position-update}
\Delta\log\pi_\theta(u_i\mid s_i)
= \eta A_i\|h_i\|^2 + \sum_{j\ne i}M_{ij} + O(\eta^2).
\end{equation}
\normalsize
The first term follows the direction of $A_i$, while the total cross-position effect can reinforce, cancel, or reverse this direction. To test the first order approximation in~\eqref{eq:cross-position-update}, we compare the predicted change $\sum_j M_{ij}$ with the measured change after one SGD step in Figure~\ref{fig:batch}(c).

\paragraph{Effects of mismatched positions.} We next examine which positions have a large effect on updates at other positions. Interaction matrices across model pairs are provided in Appendix~\ref{app:interaction-validation}. Specifically, Figure~\ref{fig:batch}(d) shows the five positions with the largest cross-position
influence in a selected batch under \texttt{Qwen3-1.7B-Base} and \texttt{
JustRL-1.5B}. Among these positions, the termination token has a large negative advantage and the largest influence on others. 

\paragraph{Takeaway.}
These results show that teacher--student mismatch at one position can affect the updates at other positions through shared model parameters. Specifically, at a position where the teacher's preferred token has near-zero student probability, a large negative advantage suppresses the student's own token but does not teach the teacher's preferred token, while the gradient at this position carries a large weight in the updates at other positions. We next examine how these realized updates affect subsequent sampling in training.

\subsection{Sampling and Trajectory Dynamics}
\label{sec:sampling-dynamics}

\begin{wrapfigure}{r}{0.60\columnwidth}
    \centering
    \captionsetup{skip=3pt}
    \captionsetup[subfigure]{skip=2pt,font=small}

    \begin{subfigure}[b]{0.47\linewidth}
        \centering
        \includegraphics[width=\linewidth]{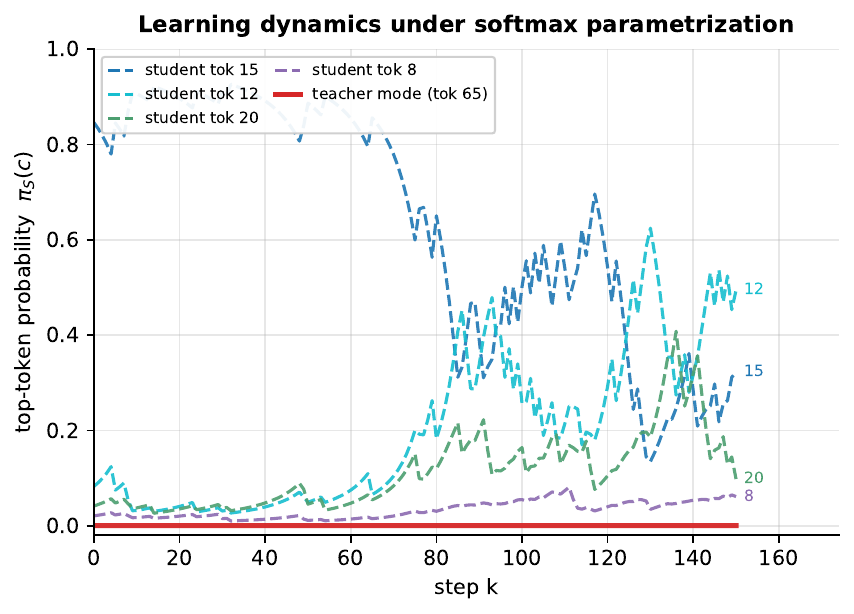}
        \caption{Categorical policy.}
        \label{fig:degeneration:toy}
    \end{subfigure}
    \hfill
    \begin{subfigure}[b]{0.47\linewidth}
        \centering
        \includegraphics[width=\linewidth]{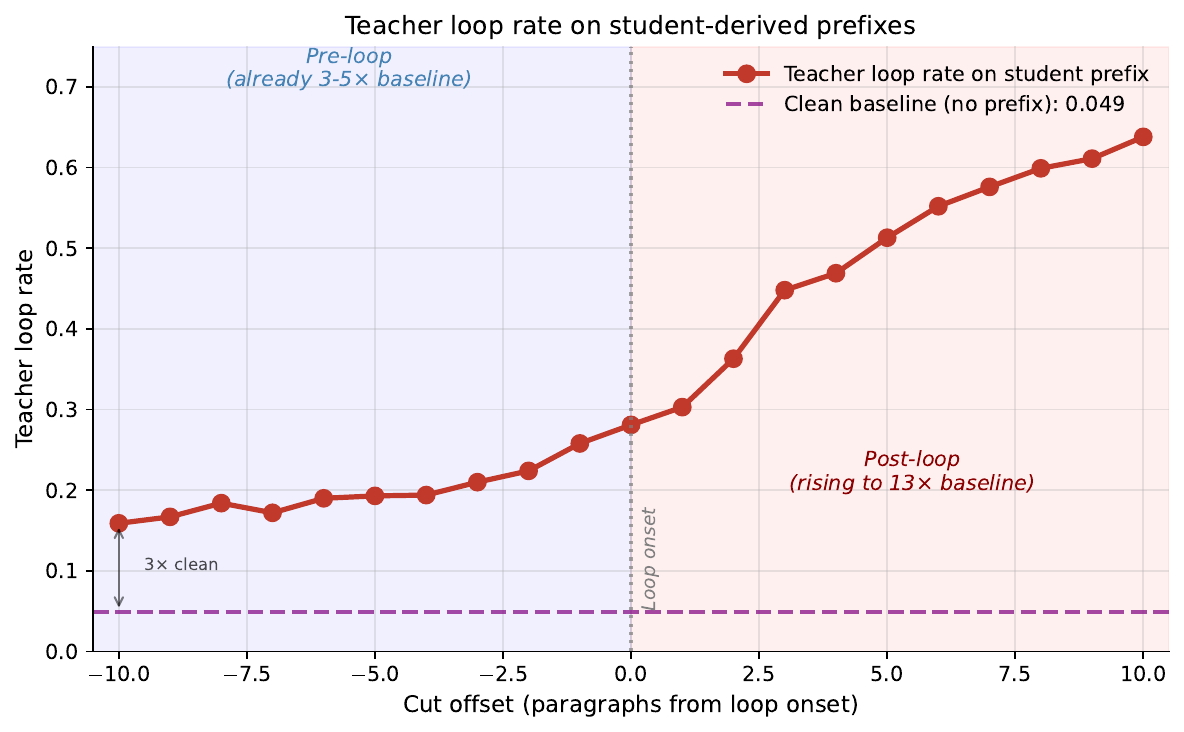}
        \caption{Teacher continuation.}
        \label{fig:teacher-loop-prefix}
    \end{subfigure}

    \caption{
    \textbf{On-policy sampling and trajectory dynamics.}
    \textbf{(a)} Token probabilities under a categorical policy.
    \textbf{(b)} Teacher loop rate when continuing student prefixes around
    the first repeated paragraph. Loop rate is the fraction of sequences in which at least five consecutive tokens are repeated at least twice.
    }
    \label{fig:degeneration}

\end{wrapfigure}

The previous sections show that under sampled-token OPD training, the student may fail to learn from the teacher at some states or even move further away from the teacher distribution. Positions with substantial teacher--student mismatch are difficult to learn from, while their gradients can still affect updates at other positions. Under on-policy sampling, these updates can also change which states the student visits next. We now examine how continued training affects subsequent sampling and generation behavior.
\vspace{-0.3cm}
\paragraph{Student mode suppression before learning the teacher's preferred token.}
We first examine what happens when the student fails to learn a teacher-preferred token. We consider two categorical policies, where they have substantial mismatch. We repeatedly apply sampled-token OPD updates and track the student's top-4 tokens and the teacher's preferred token in Figure~\ref{fig:degeneration}\subref{fig:degeneration:toy}. As training progresses, the student's mode gradually disappears as probability mass shifts to other tokens, while the probability of the teacher's preferred token remains near zero. This shows that sampled-token OPD suppresses the student's existing mode without learning the teacher's preferred token.

\paragraph{EOS-token mismatch and generation behavior.}
We next study termination token mismatch as a concrete example of how sampled-token OPD at this state affects subsequent sampling. The student (\texttt{Qwen3-1.7B-Base}) uses \texttt{<|endoftext|>} to end a response, while the teacher (\texttt{Qwen3-8B}) uses \texttt{<|im\_end|>}. During training, the student's termination token is suppressed, while the teacher's termination token remains difficult to learn. As a result, response length increases and completion rate decreases within the generation budget, while the probability mass on the overlap between the student and teacher remains high (Figure~\ref{fig:motivation}). This illustrates \emph{spurious agreement}: the student
enters longer, repetitive trajectories while maintaining
high local overlap with the teacher at the visited prefixes.
This collapse is caused by the termination-token mismatch: aligning the two termination tokens, or masking the gradient at termination tokens, removes it. (See Appendix~\ref{app:interventions} for details)

To understand why high local overlap persists between the student and teacher despite repetitive generation, we ask the teacher to continue student-generated prefixes before and after repetition begins. Figure~\ref{fig:degeneration}\subref{fig:teacher-loop-prefix} shows that the teacher rarely repeats content when generating from the original prompt. Its loop rate increases when it continues student-generated prefixes near and after the start of repetition. Thus, under this mismatch, continued training leads the student to visit states where the teacher is also more likely to repeat. At these states, the teacher cannot provide useful feedback for correcting the student's generation.
Note that not every mismatch causes this failure. It appears when the teacher places almost all of its probability on a token that the student cannot produce and that is essential to the response, such as a format token.

\section{Implications for OPD Training}

The analysis points to two ways to improve learning from teacher feedback: reduce teacher--student mismatch before OPD, and provide distribution-level feedback at states where sampled-token updates give weak correction. These motivate SFT initialization and state-dependent KL routing, respectively.

\subsection{Understanding SFT Initialization for OPD}
\label{sec:cold-start}

Prior work~\citep{li2026rethinking} has shown that SFT on high-quality data or teacher trajectories can improve subsequent OPD. Our analysis helps explain why SFT initialization helps at three levels. (1) It increases the probability of teacher-preferred
tokens, making them more likely to be sampled and updated. (2) It reduces the number of hard-to-learn tokens, and hence their weight in the updates at other positions. (3) It reduces repeated updates at mismatched states, helping preserve existing student modes and avoid states where the teacher is unreliable. Note that the role of SFT here differs from its role before RL, where it gives the model enough strong and diverse behaviors for RL to select from~\citep{limozin2026sft,kang2026quagmires,matsutani2026rl}. Before OPD, SFT brings the student closer to the specific teacher, so that it can learn stably from the teacher's feedback.

\subsection{State-Dependent KL Routing}

Although SFT can reduce teacher--student mismatch, learning from the teacher may remain difficult at some states. We use coverage in \eqref{eq:mass_support} to identify insufficient student mass on teacher-preferred tokens. To additionally measure the fraction of teacher top-$p$ mass assigned to tokens underestimated by the student, we define

\small
\begin{equation}
\label{eq:fkl-vote}
V(\theta;s_t)=
\sum_{c\in\mathcal S_p(s_t)}
\hat{\pi}_T(c\mid s_t)
\mathbf{1}\!\left[
\pi_T(c\mid s_t)/\pi_\theta(c\mid s_t)>\varepsilon
\right]
,
\end{equation}
\normalsize
where $s_t=(x,y_{<t})$ and $\hat{\pi}_T$ is the teacher distribution normalized over its top-$p$ set. The threshold $\varepsilon>1$ identifies tokens whose teacher probability exceeds $\varepsilon$ times their student probability. Thus, $V(\theta;s_t)$ measures the fraction of teacher mass within this set assigned to such tokens.

As discussed before, the teacher-preferred token with near-zero student probability receives almost no update. In contrast, the top-$K$ forward-KL term changes the logit of a token $b$, raising the same token at a rate set by its teacher probability by
\small
\begin{equation}
\Delta\theta_b \;=\; \eta\Bigl[\pi_T(b\mid s)\,\mathbf 1\{b\in\mathcal S_K(s)\}-\pi_\theta(b\mid s)\sum_{c\in\mathcal S_K(s)}\pi_T(c\mid s)\Bigr].
\end{equation}
\normalsize
We therefore apply approximate top-$K$ FKL withnout normalization when the student has low overall coverage or underestimates a substantial fraction of the teacher's probability mass:
\begin{equation}
\label{eq:routed-objective}
\begin{aligned}
r(s_t)
&=
\mathbf{1}\!\left[
V(\theta;s_t)>\varepsilon_2
\ \lor\
M_{\mathrm{top}\text{-}p}(s_t)<\varepsilon_3
\right],\\
\mathcal L_{\mathrm{route}}(\theta;s_t)
&=
(1-r(s_t))\mathcal L_{\mathrm{OPD}}(\theta;s_t)
+
r(s_t)
\sum_{c\in\mathcal S_K(s_t)}
\pi_T(c\mid s_t)
\log\frac{\pi_T(c\mid s_t)}
{\pi_\theta(c\mid s_t)}.
\end{aligned}
\end{equation}
Here, $\varepsilon_2\in[0,1]$ is the threshold on underestimated teacher mass, and $\varepsilon_3\in[0,1]$ is the minimum student coverage threshold.
$\mathcal L_{\mathrm{OPD}}(\theta;s_t)$ denotes the per-state reverse-KL objective, and $\mathcal S_K(s_t)$ is the teacher's top-$K$
set used for efficient FKL computation. Teacher probabilities are not renormalized over the top-$K$ set.
The method retains sampled-token OPD by default.

\vspace{-0.4cm}

\section{Experiment}

\subsection{Experiment Setup}

\paragraph{Models and Evaluation.} We conduct experiments on both math reasoning and code generation tasks
using three Qwen3~\citep{yang2025qwen3} models:
Qwen3-1.7B-Base, Qwen3-4B-Base,
and Qwen3-8B-Base.
We use Qwen3-8B as the teacher
model in both settings:
non-thinking mode for OPD directly from Base initialization,
and thinking mode for OPD from SFT initialization. For the math reasoning experiment, we use the DAPO dataset~\citep{yu2026dapo} for OPD training
across all student sizes. For SFT, we extract 400K prompt--response pairs from
\texttt{OpenThoughts3-1.2M}, comprising 50K distinct prompts with eight responses each.  We evaluate on mathematical reasoning benchmarks: AIME24-26~\citep{maa_aime}, and AMC23~\citep{maa2023amc}.  For the code generation task, we use the \texttt{CodeR1-12k}~\citep{ganler2025coder1} for OPD training, and evaluate on HumanEval+, and MBPP+~\citep{liu2023your},
and LiveCodeBench v6~\citep{jia2026asymmetric}. We report Avg@8, the mean accuracy across the eight responses, and Best@8, the bootstrap estimate of best-of-eight accuracy obtained by resampling with replacement. For more details, we defer to Appendix~\ref{app:experimental-details}.

\paragraph{Baselines.} We compare our method against three baselines: (i) Sampled-token OPD~\citep{lu2025onpolicydistillation}; (ii) EOPD~\citep{jin2026entropy}, which uses forward KL when the teacher has high entropy; (iii) AOPD~\citep{jia2026asymmetric}, which uses FKL when the sampled-token advantage is non-positive. 

\subsection{Main Result}

\paragraph{Mathematical Reasoning.} Table~\ref{tab:qwen_math_results} compares methods within each initialization setting, with different teacher modes across panels. From Base initialization, our method achieves the highest mean Avg@8 and Best@8 at all three scales. Relative to sampled-token OPD, mean Avg@8 improves by 2.36/2.50/1.72 percentage points and mean Best@8 by 2.60/4.26/6.24 points for 1.7B/4B/8B, respectively.

\begin{table*}[t]
\centering
\small
\resizebox{0.9\linewidth}{!}{
\setlength{\tabcolsep}{4.5pt}
\begin{tabular}{
ll|
cccc>{\columncolor{gray!15}}c|
cccc>{\columncolor{gray!15}}c|
cccc>{\columncolor{gray!15}}c|
c
}
\toprule
\multirow{2}{*}{Benchmark}
& \multirow{2}{*}{Metric}
& \multicolumn{5}{c|}{Qwen3-1.7B-Base}
& \multicolumn{5}{c|}{Qwen3-4B-Base}
& \multicolumn{5}{c|}{Qwen3-8B-Base}
& \multirow{2}{*}{\shortstack{Teacher\\Qwen3-8B}} \\
\cmidrule(lr){3-7}
\cmidrule(lr){8-12}
\cmidrule(lr){13-17}
&
& SFT & OPD & EOPD & AOPD & Ours
& SFT & OPD & EOPD & AOPD & Ours
& SFT & OPD & EOPD & AOPD & Ours
& \\
\midrule

\multicolumn{18}{c}{\textit{(a) Qwen3-8B non-thinking teacher}} \\
\midrule

\multirow{2}{*}{AMC23}
& Avg@8
& -- & 38.44 & 38.75 & 33.12 & \textbf{45.00}
& -- & 53.75 & 54.37 & 55.94 & \textbf{58.75}
& -- & 60.94 & 54.38 & 60.31 & \textbf{61.56}
& 69.38 \\
& Best@8
& -- & 64.22 & \textbf{70.68} & 58.09 & 70.65
& -- & 79.12 & 79.67 & \textbf{83.89} & 80.74
& -- & 79.90 & \textbf{84.87} & \textbf{84.87} & 83.89
& 89.58 \\

\midrule

\multirow{2}{*}{AIME24}
& Avg@8
& -- & 8.75 & 9.58 & 6.25 & \textbf{12.08}
& -- & 15.83 & 17.08 & 15.83 & \textbf{18.75}
& -- & 20.42 & 21.67 & \textbf{22.50} & \textbf{22.50}
& 25.83 \\
& Best@8
& -- & 20.68 & 18.70 & 19.45 & \textbf{24.60}
& -- & 26.25 & 26.67 & \textbf{34.10} & 32.94
& -- & 32.08 & 37.28 & 38.01 & \textbf{41.77}
& 42.84 \\

\midrule

\multirow{2}{*}{AIME25}
& Avg@8
& -- & \textbf{9.58} & 7.08 & 5.42 & \textbf{9.58}
& -- & 15.00 & 16.67 & 16.25 & \textbf{18.75}
& -- & 17.50 & \textbf{18.33} & 16.25 & 17.50
& 20.42 \\
& Best@8
& -- & 14.31 & 13.25 & 17.16 & \textbf{21.89}
& -- & 27.70 & 28.78 & 25.91 & \textbf{32.82}
& -- & 27.89 & 29.58 & 20.99 & \textbf{31.60}
& 36.71 \\

\midrule

\multirow{2}{*}{AIME26}
& Avg@8
& -- & \textbf{6.25} & 5.42 & 4.58 & 5.83
& -- & \textbf{15.42} & 13.33 & 10.83 & 13.75
& -- & 14.58 & 13.75 & 12.50 & \textbf{18.75}
& 15.83 \\
& Best@8
& -- & \textbf{19.66} & 14.80 & 12.55 & 12.12
& -- & 24.44 & 24.50 & 25.29 & \textbf{28.05}
& -- & 26.31 & 26.24 & 24.72 & \textbf{33.91}
& 33.44 \\

\midrule

\multirow{2}{*}{Mean}
& Avg@8
& -- & 15.76 & 15.21 & 12.34 & \textbf{18.12}
& -- & 25.00 & 25.36 & 24.71 & \textbf{27.50}
& -- & 28.36 & 27.03 & 27.89 & \textbf{30.08}
& 32.87 \\
& Best@8
& -- & 29.72 & 29.36 & 26.81 & \textbf{32.32}
& -- & 39.38 & 39.91 & 42.30 & \textbf{43.64}
& -- & 41.55 & 44.49 & 42.15 & \textbf{47.79}
& 50.64 \\

\midrule

\multicolumn{18}{c}{\textit{(b) Qwen3-8B thinking teacher}} \\
\midrule

\multirow{2}{*}{AMC23}
& Avg@8
& \textbf{46.56} & 44.69 & 42.50 & 45.31 & 45.94
& 74.69 & 77.81 & 77.50 & 77.81 & \textbf{80.00}
& 82.19 & 83.12 & 81.88 & 82.19 & \textbf{85.31}
& 91.56 \\
& Best@8
& \textbf{72.50} & 66.70 & 61.68 & 67.38 & 64.92
& 92.50 & 90.41 & 91.30 & 90.53 & \textbf{93.16}
& \textbf{95.00} & 93.14 & 90.79 & 91.60 & 94.06
& 99.77 \\

\midrule

\multirow{2}{*}{AIME24}
& Avg@8
& 10.83 & 13.33 & 11.67 & 12.50 & \textbf{14.58}
& 42.08 & 47.50 & 44.58 & 47.50 & \textbf{48.75}
& 46.25 & 57.92 & \textbf{60.83} & 58.33 & 59.58
& 67.50 \\
& Best@8
& 30.00 & 26.95 & 22.95 & 26.17 & \textbf{34.45}
& \textbf{73.33} & 66.27 & 67.64 & 71.16 & 70.41
& 73.33 & \textbf{79.84} & 76.22 & 75.10 & 79.77
& 78.23 \\

\midrule

\multirow{2}{*}{AIME25}
& Avg@8
& 13.75 & 16.67 & 13.75 & \textbf{18.75} & 17.08
& 34.17 & 37.08 & \textbf{42.08} & 36.25 & 37.50
& 37.08 & 40.42 & 40.83 & 40.83 & \textbf{41.67}
& 53.33 \\
& Best@8
& 23.33 & 28.44 & 27.80 & \textbf{29.81} & 28.34
& \textbf{60.00} & 57.32 & 58.77 & 56.92 & 57.25
& 56.67 & 61.30 & 60.23 & 59.05 & \textbf{62.31}
& 73.68 \\

\midrule

\multirow{2}{*}{AIME26}
& Avg@8
& 9.17 & 8.75 & 11.25 & \textbf{15.42} & 14.17
& 37.08 & 40.83 & 42.08 & 41.25 & \textbf{42.50}
& 46.25 & 50.83 & 47.08 & 48.33 & \textbf{52.08}
& 55.83 \\
& Best@8
& 26.67 & 18.43 & 28.01 & \textbf{29.60} & 26.12
& \textbf{60.00} & 54.76 & 52.94 & 55.57 & 57.15
& 66.67 & 69.39 & 62.56 & \textbf{72.07} & 71.56
& 71.37 \\

\midrule

\multirow{2}{*}{Mean}
& Avg@8
& 20.08 & 20.86 & 19.79 & \textbf{23.00} & 22.94
& 47.01 & 50.81 & 51.56 & 50.70 & \textbf{52.19}
& 52.94 & 58.07 & 57.66 & 57.42 & \textbf{59.66}
& 67.06 \\
& Best@8
& 38.13 & 35.13 & 35.11 & 38.24 & \textbf{38.46}
& \textbf{71.46} & 67.19 & 67.66 & 68.55 & 69.49
& 72.92 & 75.92 & 72.45 & 74.46 & \textbf{76.93}
& 80.76 \\

\bottomrule
\end{tabular}
}
\vspace{-0.1cm}
\caption{
Math reasoning results with Qwen3-8B as teacher. (a) Base initialization, non-thinking teacher. (b) Shared SFT initialization per model size, thinking teacher. Our routing settings are $p=0.9$, $K=16$, $\varepsilon_1=6$, $\varepsilon_2=0.3$, and $\varepsilon_3=0.2$. Bold indicates the best result.
}
\label{tab:qwen_math_results}
\end{table*}

With SFT initialization, the corresponding gains are 2.08/1.38/1.59 points in mean Avg@8 and 3.33/2.30/1.01 points in mean Best@8. Our method achieves the highest mean Best@8 among OPD methods at all three scales, and the highest mean Avg@8 at 4B and 8B, with a comparable result at 1.7B. For the 4B student, all OPD variants improve Avg@8 but reduce Best@8 relative to the SFT checkpoint. This pattern is consistent with a trade-off previously discussed for on-policy RL, which concentrate probability mass on fewer solution modes, improving Pass@1 while reducing Pass@$k$~\citep{yue2026does,matsutani2026rl}. 

\begin{figure}[h]
    \centering
    \includegraphics[width=0.94\linewidth]{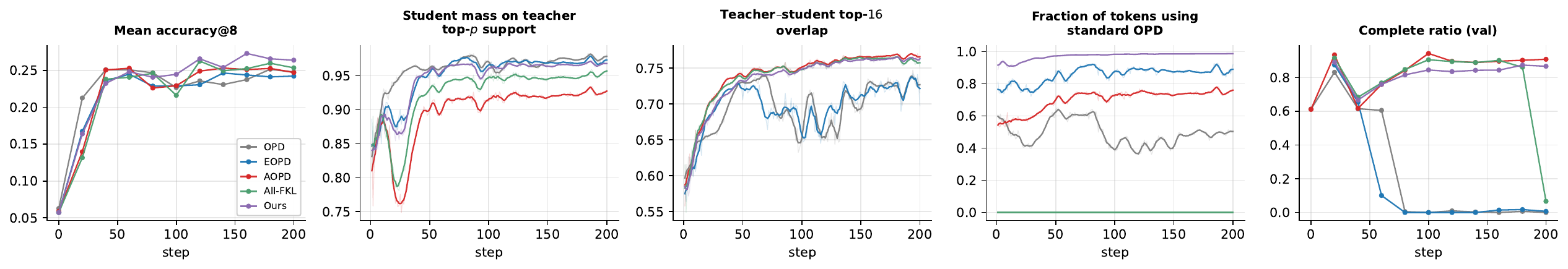}
    \vspace{-0.2cm}
    \caption{Training dynamics for Qwen3-4B-Base with the Qwen3-8B non-thinking teacher.}
    \label{fig:training_dynamics_4b}
\end{figure}

Figure~\ref{fig:training_dynamics_4b} shows the training dynamics curve of different methods, includes All-FKL, which applies top-$K$ FKL at every position. OPD and EOPD exhibit \emph{spurious agreement}: top-$16$ overlap and teacher-support coverage remain high despite collapsing completion rates. Our method maintains stable completion and the highest coverage of the teacher's top-$p$ support, achieving the best performance while using FKL at the fewest positions among the adaptive methods. We also compare against All-FKL, which applies FKL at every position but does not achieve better performance. In terms of cost, methods that use forward KL require the teacher's top-$K$ log-probabilities instead of only the log-probability of the sampled token. For our method with $K=16$, the time per training step increases by 2.5--6.8\% over sampled-token OPD across the three student sizes, and the peak memory of a training step is unchanged (See Appendix~\ref{app:overhead}) for the details.

\paragraph{Code Generation.} 
Table~\ref{tab:qwen_code_results} reports results from base initialization with the non-thinking teacher. Our method achieves the highest mean Avg@8 at all three scales, and the highest mean Best@8 at 1.7B and 8B. Relative to sampled-token OPD, mean Avg@8 improves by 2.92/3.19/3.44 percentage points and mean Best@8 by 2.08/1.27/2.88 points for 1.7B/4B/8B, respectively. The gain comes mainly from LiveCodeBench, where Avg@8 improves by 6.00/10.93/8.14 points, while results on HumanEval+ and MBPP+ are close to those of sampled-token OPD. For the 4B student, AOPD has a higher mean Best@8.
\begin{table*}[t]
\centering
\small
\resizebox{0.9\linewidth}{!}{
\setlength{\tabcolsep}{4.5pt}
\begin{tabular}{
ll|
ccc>{\columncolor{gray!15}}c|
ccc>{\columncolor{gray!15}}c|
ccc>{\columncolor{gray!15}}c
}
\toprule
\multirow{2}{*}{Benchmark}
& \multirow{2}{*}{Metric}
& \multicolumn{4}{c|}{Qwen3-1.7B-Base}
& \multicolumn{4}{c|}{Qwen3-4B-Base}
& \multicolumn{4}{c}{Qwen3-8B-Base} \\
\cmidrule(lr){3-6}
\cmidrule(lr){7-10}
\cmidrule(lr){11-14}
&
& OPD & EOPD & AOPD & Ours
& OPD & EOPD & AOPD & Ours
& OPD & EOPD & AOPD & Ours \\
\midrule

\multirow{2}{*}{HumanEval+}
& Avg@8
& 64.71 & 65.17 & 64.64 & \textbf{67.07}
& \textbf{79.27} & 78.43 & 73.09 & 79.19
& 78.73 & 79.27 & 79.65 & \textbf{80.64} \\
& Best@8
& 77.26 & 77.39 & \textbf{78.27} & 76.76
& 85.97 & \textbf{88.44} & 87.82 & 86.64
& 82.21 & 84.85 & 86.32 & \textbf{87.28} \\

\midrule

\multirow{2}{*}{MBPP+}
& Avg@8
& 55.59 & 55.19 & 55.56 & \textbf{55.99}
& \textbf{68.39} & 66.57 & 62.90 & 67.10
& 72.45 & 72.39 & 70.27 & \textbf{72.72} \\
& Best@8
& 66.10 & 66.42 & 66.60 & \textbf{68.60}
& 74.11 & 73.57 & \textbf{77.70} & 73.83
& 77.77 & 77.98 & 78.97 & \textbf{79.22} \\

\midrule

\multirow{2}{*}{LiveCodeBench v6}
& Avg@8
& 8.93 & 10.36 & 12.07 & \textbf{14.93}
& 10.00 & 9.36 & 19.07 & \textbf{20.93}
& 15.50 & 14.07 & 21.64 & \textbf{23.64} \\
& Best@8
& 17.49 & 18.49 & 19.58 & \textbf{21.75}
& 24.72 & 21.54 & 27.25 & \textbf{28.14}
& 29.68 & 27.98 & 31.40 & \textbf{31.81} \\

\midrule

\multirow{2}{*}{Mean}
& Avg@8
& 43.08 & 43.57 & 44.09 & \textbf{46.00}
& 52.55 & 51.45 & 51.69 & \textbf{55.74}
& 55.56 & 55.24 & 57.19 & \textbf{59.00} \\
& Best@8
& 53.62 & 54.10 & 54.81 & \textbf{55.70}
& 61.60 & 61.18 & \textbf{64.26} & 62.87
& 63.22 & 63.60 & 65.56 & \textbf{66.10} \\

\bottomrule
\end{tabular}
}
\vspace{-0.1cm}
\caption{
Code generation results with Qwen3-8B (non-thinking) as teacher, from Base initialization, trained on \texttt{CodeR1-12k}. We report the checkpoint at step 200 for all methods. Our routing settings are $p=0.9$, $K=16$, $\varepsilon_1=6$, $\varepsilon_2=0.3$, and $\varepsilon_3=0.2$. Bold indicates the best result.
}
\label{tab:qwen_code_results}
\end{table*}

\paragraph{Ablation Study.} Appendix~\ref{app:ablation} presents ablation studies examining sensitivity of our algorithm.

\vspace{-0.2cm}
\section{Conclusion}
Our analysis explains why a strong teacher does not always lead to effective learning under sampled-token OPD. Local updates may provide little correction to teacher-preferred tokens, gradients from other positions can alter update directions, and continued training can lead to repetitive trajectories with high teacher--student overlap. These findings help explain the benefits of SFT initialization and motivate state-dependent KL routing, which improves performance across Qwen3 model scales on mathematical reasoning benchmarks.

\newpage

\appendix

\subsection*{AI use statement}

We used generative AI tools only to assist with writing and plotting. The authors reviewed all AI-assisted outputs and take full responsibility for the content of the manuscript and all reported results.

\subsection*{Ethics statement}

This work studies the learning dynamics of on-policy
distillation and methods for improving student learning
from teacher feedback. Although our experiments focus
on mathematical reasoning, the proposed method could
also be applied to other language-model tasks, where
improved model capabilities may introduce risks of
misuse.

\subsection*{Reproducibility statement}

Our experiments use public datasets and models,
with training implemented using the open-source verl framework.
The appendix~\ref{app:experimental-details} documents the training hyperparameters,
evaluation protocols, baseline implementations, and
additional experimental results to support reproducibility.

\bibliography{iclr2027_conference}

\begin{thebibliography}{33}
\providecommand{\natexlab}[1]{#1}
\providecommand{\url}[1]{\texttt{#1}}
\expandafter\ifx\csname urlstyle\endcsname\relax
  \providecommand{\doi}[1]{doi: #1}\else
  \providecommand{\doi}{doi: \begingroup \urlstyle{rm}\Url}\fi

\bibitem[Agarwal et~al.(2024)Agarwal, Vieillard, Zhou, Stanczyk, Ramos~Garea,
  Geist, and Bachem]{agarwal2024policy}
Rishabh Agarwal, Nino Vieillard, Yongchao Zhou, Piotr Stanczyk, Sabela
  Ramos~Garea, Matthieu Geist, and Olivier Bachem.
\newblock On-policy distillation of language models: Learning from
  self-generated mistakes.
\newblock In \emph{International Conference on Learning Representations},
  volume 2024, pp.\  21246--21263, 2024.

\bibitem[Armandpour et~al.(2026)Armandpour, Ilhan, Harrison, Jaiswal, Hoang,
  Faghri, Zhang, Cho, and Farajtabar]{armandpour2026unmasking}
Mohammadreza Armandpour, Fatih Ilhan, David Harrison, Ajay Jaiswal, Duc~NM
  Hoang, Fartash Faghri, Yizhe Zhang, Minsik Cho, and Mehrdad Farajtabar.
\newblock Unmasking on-policy distillation: Where it helps, where it hurts, and
  why.
\newblock \emph{arXiv preprint arXiv:2605.10889}, 2026.

\bibitem[Bengio et~al.(2015)Bengio, Vinyals, Jaitly, and
  Shazeer]{bengio2015scheduled}
Samy Bengio, Oriol Vinyals, Navdeep Jaitly, and Noam Shazeer.
\newblock Scheduled sampling for sequence prediction with recurrent neural
  networks.
\newblock \emph{Advances in neural information processing systems}, 28, 2015.

\bibitem[Fu et~al.(2026)Fu, Huang, Jiang, Liu, Jiang, Zhu, and
  Zhao]{fu2026revisiting}
Yuqian Fu, Haohuan Huang, Kaiwen Jiang, Jiacai Liu, Zhuo Jiang, Yuanheng Zhu,
  and Dongbin Zhao.
\newblock Revisiting on-policy distillation: Empirical failure modes and simple
  fixes.
\newblock \emph{arXiv preprint arXiv:2603.25562}, 2026.

\bibitem[{ganler}(2025)]{ganler2025coder1}
{ganler}.
\newblock {CodeR1-12K}.
\newblock Hugging Face dataset repository, 2025.
\newblock URL \url{https://huggingface.co/datasets/ganler/code-r1-12k}.
\newblock Revision 2a0d26ddd96adc28138ad9cd2c371c27ba5614ac.

\bibitem[Gu et~al.(2024)Gu, Dong, Wei, and Huang]{gu2024minillm}
Yuxian Gu, Li~Dong, Furu Wei, and Minlie Huang.
\newblock Minillm: Knowledge distillation of large language models.
\newblock In \emph{International Conference on Learning Representations},
  volume 2024, pp.\  32694--32717, 2024.

\bibitem[Hinton et~al.(2015)Hinton, Vinyals, and Dean]{hinton2015distilling}
Geoffrey Hinton, Oriol Vinyals, and Jeff Dean.
\newblock Distilling the knowledge in a neural network.
\newblock \emph{arXiv preprint arXiv:1503.02531}, 2015.

\bibitem[Jia et~al.(2026)Jia, Yang, Ma, Lian, Zhang, Zhang, Zeng, Cai, and
  Sun]{jia2026asymmetric}
Nan Jia, Haojin Yang, Xing Ma, Jiesong Lian, Shuailiang Zhang, Weipeng Zhang,
  Ke~Zeng, Xunliang Cai, and Zequn Sun.
\newblock Asymmetric on-policy distillation: Bridging exploitation and
  imitation at the token level.
\newblock \emph{arXiv preprint arXiv:2605.06387}, 2026.

\bibitem[Jin et~al.(2026)Jin, Min, Yang, Wei, Zhou, Kadhe, Baracaldo, and
  Lee]{jin2026entropy}
Woogyeol Jin, Taywon Min, Yongjin Yang, Dennis Wei, Yi~Zhou, Swanand~Ravindra
  Kadhe, Nathalie Baracaldo, and Kimin Lee.
\newblock Entropy-aware on-policy distillation of language models.
\newblock \emph{arXiv preprint arXiv:2603.07079}, 2026.

\bibitem[Kang et~al.(2026)Kang, Kuchnik, Padthe, Vlastelica, Jia, Wu, and
  Ardalani]{kang2026quagmires}
Feiyang Kang, Michael Kuchnik, Karthik Padthe, Marin Vlastelica, Ruoxi Jia,
  Carole-Jean Wu, and Newsha Ardalani.
\newblock Quagmires in sft-rl post-training: When high sft scores mislead and
  what to use instead.
\newblock In \emph{International Conference on Learning Representations},
  volume 2026, pp.\  56876--56918, 2026.

\bibitem[Kim \& Rush(2016)Kim and Rush]{kim2016sequence}
Yoon Kim and Alexander~M Rush.
\newblock Sequence-level knowledge distillation.
\newblock In \emph{Proceedings of the 2016 conference on empirical methods in
  natural language processing}, pp.\  1317--1327, 2016.

\bibitem[Ko et~al.(2026)Ko, Abdali, Kim, Chen, and Cameron]{ko2026scaling}
Jongwoo Ko, Sara Abdali, Young~Jin Kim, Tianyi Chen, and Pashmina Cameron.
\newblock Scaling reasoning efficiently via relaxed on-policy distillation.
\newblock \emph{arXiv preprint arXiv:2603.11137}, 2026.

\bibitem[Li et~al.(2026)Li, Zuo, He, Zhang, Xiao, Qian, Yu, Gao, Yang, Liu,
  et~al.]{li2026rethinking}
Yaxuan Li, Yuxin Zuo, Bingxiang He, Jinqian Zhang, Chaojun Xiao, Cheng Qian,
  Tianyu Yu, Huan-ang Gao, Wenkai Yang, Zhiyuan Liu, et~al.
\newblock Rethinking on-policy distillation of large language models:
  Phenomenology, mechanism, and recipe.
\newblock \emph{arXiv preprint arXiv:2604.13016}, 2026.

\bibitem[Limozin et~al.(2026)Limozin, Durech, Hoefler, Schlag, and
  Pyatkin]{limozin2026sft}
Alexis Limozin, Eduard Durech, Torsten Hoefler, Imanol Schlag, and Valentina
  Pyatkin.
\newblock Sft-then-rl outperforms mixed-policy methods for llm reasoning.
\newblock \emph{arXiv preprint arXiv:2604.23747}, 2026.

\bibitem[Liu et~al.(2023)Liu, Xia, Wang, and Zhang]{liu2023your}
Jiawei Liu, Chunqiu~Steven Xia, Yuyao Wang, and Lingming Zhang.
\newblock Is your code generated by chatgpt really correct? rigorous evaluation
  of large language models for code generation.
\newblock \emph{Advances in neural information processing systems},
  36:\penalty0 21558--21572, 2023.

\bibitem[Lu \& Lab(2025)Lu and Lab]{lu2025onpolicydistillation}
Kevin Lu and Thinking~Machines Lab.
\newblock On-policy distillation.
\newblock \emph{Thinking Machines Lab: Connectionism}, 2025.
\newblock \doi{10.64434/tml.20251026}.
\newblock https://thinkingmachines.ai/blog/on-policy-distillation.

\bibitem[Luo et~al.(2026)Luo, Chuang, Wang, Xu, Han, Zhang, and
  Braverman]{luo2026demystifying}
Feng Luo, Yu-Neng Chuang, Guanchu Wang, Zicheng Xu, Xiaotian Han, Tianyi Zhang,
  and Vladimir Braverman.
\newblock Demystifying opd: Length inflation and stabilization strategies for
  large language models.
\newblock \emph{arXiv preprint arXiv:2604.08527}, 2026.

\bibitem[Matsutani et~al.(2026)Matsutani, Takashiro, Minegishi, Kojima,
  Iwasawa, and Matsuo]{matsutani2026rl}
Kohsei Matsutani, Shota Takashiro, Gouki Minegishi, Takeshi Kojima, Yusuke
  Iwasawa, and Yutaka Matsuo.
\newblock Rl squeezes, sft expands: A comparative study of reasoning llms.
\newblock In \emph{International Conference on Learning Representations},
  volume 2026, pp.\  141482--141526, 2026.

\bibitem[{MMA}()]{maa_aime}
{MMA}.
\newblock {MAA} invitational competitions.
\newblock \url{https://maa.org/maa-invitational-competitions/}.
\newblock Accessed: September 8, 2026.

\bibitem[{MMA}(2023)]{maa2023amc}
{MMA}.
\newblock 2023 american mathematics competitions.
\newblock \url{https://maa.org/student-programs/amc/}, 2023.

\bibitem[Ren \& Sutherland(2025)Ren and Sutherland]{ren2025learning}
Yi~Ren and Danica Sutherland.
\newblock Learning dynamics of llm finetuning.
\newblock In \emph{International Conference on Learning Representations},
  volume 2025, pp.\  70523--70563, 2025.

\bibitem[Ross et~al.(2011)Ross, Gordon, and Bagnell]{ross2011reduction}
St{\'e}phane Ross, Geoffrey Gordon, and Drew Bagnell.
\newblock A reduction of imitation learning and structured prediction to
  no-regret online learning.
\newblock In \emph{Proceedings of the fourteenth international conference on
  artificial intelligence and statistics}, pp.\  627--635. JMLR Workshop and
  Conference Proceedings, 2011.

\bibitem[Shao et~al.(2024)Shao, Wang, Zhu, Xu, Song, Bi, Zhang, Zhang, Li, Wu,
  et~al.]{shao2024deepseekmath}
Zhihong Shao, Peiyi Wang, Qihao Zhu, Runxin Xu, Junxiao Song, Xiao Bi, Haowei
  Zhang, Mingchuan Zhang, YK~Li, Yang Wu, et~al.
\newblock Deepseekmath: Pushing the limits of mathematical reasoning in open
  language models.
\newblock \emph{arXiv preprint arXiv:2402.03300}, 2024.

\bibitem[Wang et~al.(2026)Wang, Lu, Gu, Wang, Yang, Yan, Xie, Wu, and
  Yang]{wang2026not}
Yuanyi Wang, Su~Lu, Yanggan Gu, Pengkai Wang, Yifan Yang, Zhaoyi Yan, Congkai
  Xie, Jianmin Wu, and Hongxia Yang.
\newblock Not all disagreement is learnable: Token teachability in on-policy
  distillation.
\newblock \emph{arXiv preprint arXiv:2605.26844}, 2026.

\bibitem[Xiao et~al.(2026)Xiao, Xia, Yang, Gao, Shen, Zhang, He, Lou, Luo,
  Wang, et~al.]{xiao2026mimo}
Bangjun Xiao, Bingquan Xia, Bo~Yang, Bofei Gao, Bowen Shen, Chen Zhang,
  Chenhong He, Chiheng Lou, Fuli Luo, Gang Wang, et~al.
\newblock Mimo-v2-flash technical report.
\newblock \emph{arXiv preprint arXiv:2601.02780}, 2026.

\bibitem[Xing et~al.(2026)Xing, Wang, Gao, Li, and Tang]{xing2026trust}
Xingrun Xing, Haoqing Wang, Boyan Gao, Ziheng Li, and Yehui Tang.
\newblock Trust region on-policy distillation.
\newblock \emph{arXiv preprint arXiv:2606.01249}, 2026.

\bibitem[Xu et~al.(2026{\natexlab{a}})Xu, Lin, Xue, Wang, Xu, Wu, Zhang, Lin,
  Dong, Ling, et~al.]{xu2026deepseek}
Anyi Xu, Bangcai Lin, Bing Xue, Bingxuan Wang, Bingzheng Xu, Bochao Wu, Bowei
  Zhang, Chaofan Lin, Chen Dong, Chenchen Ling, et~al.
\newblock Deepseek-v4: Towards highly efficient million-token context
  intelligence.
\newblock \emph{arXiv preprint arXiv:2606.19348}, 2026{\natexlab{a}}.

\bibitem[Xu et~al.(2026{\natexlab{b}})Xu, Sang, Zhou, He, Wang, and
  Geramifard]{xu2026tip}
Yuanda Xu, Hejian Sang, Zhengze Zhou, Ran He, Zhipeng Wang, and Alborz
  Geramifard.
\newblock Tip: Token importance in on-policy distillation.
\newblock \emph{arXiv preprint arXiv:2604.14084}, 2026{\natexlab{b}}.

\bibitem[Yang et~al.(2025)Yang, Li, Yang, Zhang, Hui, Zheng, Yu, Gao, Huang,
  Lv, et~al.]{yang2025qwen3}
An~Yang, Anfeng Li, Baosong Yang, Beichen Zhang, Binyuan Hui, Bo~Zheng, Bowen
  Yu, Chang Gao, Chengen Huang, Chenxu Lv, et~al.
\newblock Qwen3 technical report.
\newblock \emph{arXiv preprint arXiv:2505.09388}, 2025.

\bibitem[Yu et~al.(2026)Yu, Zhang, Zhu, Yuan, Zuo, Yue, Dai, Fan, Liu, Liu,
  et~al.]{yu2026dapo}
Qiying Yu, Zheng Zhang, Ruofei Zhu, Yufeng Yuan, Xiaochen Zuo, Yu~Yue, Weinan
  Dai, Tiantian Fan, Gaohong Liu, Lingjun Liu, et~al.
\newblock Dapo: An open-source llm reinforcement learning system at scale.
\newblock \emph{Advances in Neural Information Processing Systems},
  38:\penalty0 113222--113244, 2026.

\bibitem[Yue et~al.(2026)Yue, Chen, Lu, Zhao, Wang, Song, and
  Huang]{yue2026does}
Yang Yue, Zhiqi Chen, Rui Lu, Andrew Zhao, Zhaokai Wang, Shiji Song, and Gao
  Huang.
\newblock Does reinforcement learning really incentivize reasoning capacity in
  llms beyond the base model?
\newblock \emph{Advances in Neural Information Processing Systems},
  38:\penalty0 57654--57689, 2026.

\bibitem[Zheng et~al.(2024)Zheng, Zhang, Zhang, Ye, and
  Luo]{zheng2024llamafactory}
Yaowei Zheng, Richong Zhang, Junhao Zhang, Yanhan Ye, and Zheyan Luo.
\newblock Llamafactory: Unified efficient fine-tuning of 100+ language models.
\newblock In \emph{Proceedings of the 62nd annual meeting of the association
  for computational linguistics (volume 3: system demonstrations)}, pp.\
  400--410, 2024.

\bibitem[Zhu et~al.(2026)Zhu, Ye, Lu, Shi, and Liu]{zhu2026many}
Siqi Zhu, Xuyan Ye, Hongyu Lu, Weiye Shi, and Ge~Liu.
\newblock The many faces of on-policy distillation: Pitfalls, mechanisms, and
  fixes.
\newblock \emph{arXiv preprint arXiv:2605.11182}, 2026.

\end{thebibliography}
\bibliographystyle{iclr2027_conference}

\newpage

\appendix

\section{Related Work}

\paragraph{Knowledge Distillation.}
Knowledge distillation trains a student using predictions
from a teacher~\citep{hinton2015distilling}.
For sequence generation, \citet{kim2016sequence} introduced
sequence-level distillation using teacher-generated responses.
The student is trained on teacher-generated responses but generates its own responses at inference time. This mismatch leads to exposure bias~\citep{bengio2015scheduled}, with errors accumulating over long sequences.
This motivates on-policy distillation, which
provides teacher feedback on student-generated
prefixes~\citep{agarwal2024policy}.

\paragraph{On-Policy Distillation.}
MiniLLM~\citep{gu2024minillm} develops an on-policy
optimization approach for reverse-KL distillation.
GKD~\citep{agarwal2024policy} supports different divergence
objectives and mixtures of fixed data and student-generated
sequences.
Recent methods adapt supervision at individual token positions.
EOPD~\citep{jin2026entropy} adds FKL where teacher entropy
is high, while AOPD~\citep{jia2026asymmetric} uses FKL
when the sampled-token advantage is non-positive.
TrOPD~\citep{xing2026trust} uses the sampled-token
teacher--student probability ratio to define a trust region,
applies FKL to outliers, and adds teacher-prefix guidance.
Other methods adjust or select reverse-KL supervision.
REOPOLD~\citep{ko2026scaling} combines reward clipping
with entropy-based token selection.
TIP~\citep{xu2026tip} studies token importance through
student entropy and teacher--student divergence, while
Teachability-Aware OPD~\citep{wang2026not}
prioritizes disagreement within the student's current support.
Our analysis shows why teacher entropy and sampled advantage
alone can miss substantial mismatch: a confident teacher
may prefer tokens the student rarely samples, and a positive
advantage on the sampled token does not rule out mismatch
on other tokens.
We therefore use teacher-support coverage and underestimation
to identify positions where sampled-token reverse-KL updates
provide weak correction, and apply FKL to directly supervise
teacher-preferred alternatives.

\paragraph{Learning Dynamics of LLM Post-Training.}
\citet{ren2025learning} provide a framework and analyze how fine-tuning updates affect predictions across examples and responses under different post-training algorithms.
For OPD, \citet{li2026rethinking} study teacher--student
compatibility, alignment among high-probability tokens,
and the benefits of teacher-trajectory cold start.
They also examine declining teacher supervision quality
on longer student prefixes and suggest that cross-position
gradient cancellation may limit learning.
We build on these perspectives by tracing sampled-token OPD
updates through local distributions, generated trajectories,
and shared-parameter optimization.
Our analysis connects mode suppression and teacher-mode
acquisition, identifies spurious agreement on repetitive
trajectories, and directly measures cross-position
contributions to realized updates.
These results help explain the benefits of SFT initialization and guide the choice of positions at which to apply FKL.

\section{Proof}

\subsection{Relation Between Eq.~\eqref{eq:prelim} and Eq.~\eqref{eq:prelim:opd-grad}}
\label{app:objective-relation}

Here we show how the sampled-token update in
Eq.~\eqref{eq:prelim:opd-grad} relates to the sequence-level objective in
Eq.~\eqref{eq:prelim}. For simplicity, we treat $T$ as a fixed horizon, write $d_t$ for the
distribution of prefixes $s_t=(x,y_{<t})$ under the student, and let
$\mathrm{KL}_t(s_t) := D_{\mathrm{KL}}\bigl(\pi_\theta(\cdot|s_t)\,\|\,\pi_T(\cdot|s_t)\bigr)$.

We start from the exact gradient. Since
$\mathbb E_{y_t\sim\pi_\theta(\cdot|s_t)}[-A_t]=\mathrm{KL}_t(s_t)$, the
objective can be written as
$\mathcal L_{\mathrm{OPD}}(\theta)=\mathbb E_{y\sim\pi_\theta}\bigl[-\sum_{t'}A_{t'}\bigr]$.
Differentiating gives following relation:
\begin{equation}
\label{eq:grad-two-terms}
\nabla_\theta\mathcal L_{\mathrm{OPD}}(\theta)
\;=\; -\,\mathbb E_{y\sim\pi_\theta}\Bigl[\Bigl(\sum_{t'=1}^{T}A_{t'}\Bigr)\sum_{t=1}^{T}\nabla_\theta\log\pi_\theta(y_t|s_t)\Bigr]
\;+\; \mathbb E_{y\sim\pi_\theta}\Bigl[\sum_{t=1}^{T}\nabla_\theta\log\pi_\theta(y_t|s_t)\Bigr],
\end{equation}
where the second term arises because $A_{t'}$ itself depends on $\theta$.
Since the score function has zero mean at every state, thus the second term is zero.
In addition, the same fact removes the cross terms with $t'<t$ in the first term because
$A_{t'}$ is  fixed once $s_t$ is given. Therefore, the gradient of sequence-level objective can be written as
\begin{equation}
\label{eq:full-grad}
\nabla_\theta\mathcal L_{\mathrm{OPD}}(\theta)
\;=\; -\,\mathbb E_{y\sim\pi_\theta}\Bigl[\sum_{t=1}^{T}\Bigl(\sum_{t'\ge t}A_{t'}\Bigr)\nabla_\theta\log\pi_\theta(y_t|s_t)\Bigr],
\end{equation}
in which each token is weighted by its reward-to-go term.

In \citet{lu2025onpolicydistillation}, Eq.~\eqref{eq:prelim:opd-grad} keeps only the $t'=t$ part of this sum. Note that at a fixed state $s_t$,
\begin{equation}
\label{eq:per-state-grad}
\nabla_\theta\mathrm{KL}_t(s_t)
\;=\; -\,\mathbb E_{y_t\sim\pi_\theta(\cdot|s_t)}\bigl[A_t\,\nabla_\theta\log\pi_\theta(y_t|s_t)\bigr].
\end{equation}
Therefore, the $t'=t$ part is the gradient of the per-state reverse KL with the prefix distribution held fixed. The remaining $t'>t$ part accounts for the dependence of that distribution on
$\theta$. 

In words, Eq.~\eqref{eq:prelim:opd-grad} is the policy gradient of
Eq.~\eqref{eq:prelim} with the reward-to-go truncated to the immediate term,
which is the same as fixing the distribution of student-visited prefixes. The
sampled-token estimator is therefore unbiased for
Eq.~\eqref{eq:prelim:opd-grad}, but not for the exact gradient of
Eq.~\eqref{eq:prelim}.

\subsection{Proof of Proposition~\ref{prop:per-token}}
\label{app:proof-prop1}

Fix a state $s\in\mathcal S$. Write $\theta_b^k := \theta_{s,b}^{k}$, and define
\[
\Delta \theta_b \;:=\; \theta_b^{k+1}-\theta_b^k,\qquad
\Delta\pi(c|s) \;:=\; \pi_\theta^{k+1}(c|s)-\pi_\theta^k(c|s).
\]

By assumption, the logits $\{\theta_b\}_{b\in\mathcal V}$ at state $s$ are
parameterized independently of those at any other state, so we may
treat them as the free variables. Taylor-expanding the softmax around
$\theta^k$ to first order in $\Delta\theta$,
\begin{equation}
\label{eq:taylor}
\Delta\pi(c|s) \;=\; \sum_{b\in\mathcal V}\frac{\partial \pi_\theta^k(c|s)}{\partial \theta_b}\,\Delta \theta_b \;+\; O(\|\Delta\theta\|^2).
\end{equation}
The softmax Jacobian is
\begin{equation}
\label{eq:softmax-jacobian}
\frac{\partial \pi_\theta^k(c|s)}{\partial \theta_b} \;=\; \pi_\theta^k(c|s)\,\bigl(\mathbf 1\{c=b\}-\pi_\theta^k(b|s)\bigr).
\end{equation}

At the fixed state $s$, define the token-level advantage
\begin{equation}
\label{eq:advantage}
A(s,a) \;:=\; \log\pi_T(a|s)-\log\pi_\theta^k(a|s)
\end{equation}
and hold it fixed at iteration $k$ to form the local surrogate objective
\[
J(\theta) \;=\; \mathbb E_{a\sim\pi_\theta(\cdot|s)}\bigl[A(s,a)\bigr].
\]
At $\theta=\theta^k$, the gradient of $J(\theta)$ equals the gradient
of the negative reverse KL between $\pi_\theta(\cdot|s)$ and
$\pi_T(\cdot|s)$; the two objectives are not identical for general $\theta$.
Drawing one sample $u\sim\pi_\theta^k(\cdot|s)$ and treating $A(s,u)$
as a stop-gradient scalar, the unbiased policy-gradient estimator of
$\left.\nabla_\theta J(\theta)\right|_{\theta=\theta^k}$ has logit-coordinate components
\begin{equation}
\label{eq:logit-grad}
\frac{\partial \widehat{J}}{\partial \theta_b} \;=\; A(s,u)\,\frac{\partial \log\pi_\theta^k(u|s)}{\partial \theta_b} \;=\; A(s,u)\,\bigl(\mathbf 1\{u=b\}-\pi_\theta^k(b|s)\bigr),
\end{equation}
where the second equality uses the standard log-softmax derivative
$\partial_{\theta_b}\log\pi_\theta^k(u|s) = \mathbf 1\{u=b\}-\pi_\theta^k(b|s)$.
One gradient ascent step with learning rate $\eta$ therefore yields
\begin{equation}
\label{eq:logit-update}
\Delta \theta_b \;=\; \eta\,A(s,u)\,\bigl(\mathbf 1\{u=b\}-\pi_\theta^k(b|s)\bigr).
\end{equation}

\medskip

Substituting~\eqref{eq:softmax-jacobian} and~\eqref{eq:logit-update}
into~\eqref{eq:taylor}, the first-order term at token $c$ is
\[
\sum_{b\in\mathcal V}\frac{\partial \pi_\theta^k(c|s)}{\partial \theta_b}\,\Delta \theta_b
\;=\; \eta\,A(s,u)\,\pi_\theta^k(c|s)\sum_{b\in\mathcal V}\bigl(\mathbf 1\{c=b\}-\pi_\theta^k(b|s)\bigr)\bigl(\mathbf 1\{u=b\}-\pi_\theta^k(b|s)\bigr).
\]
The remainder term in~\eqref{eq:taylor} is $O(\|\Delta\theta\|^2)=
O(\eta^2 A(s,u)^2)$, which we absorb into $O(\eta^2)$ for fixed
$A(s,u)$. Defining
\begin{equation}
\label{eq:Bdef}
B(c,u;s) \;:=\; \sum_{b\in\mathcal V}\bigl(\mathbf 1\{c=b\}-\pi_\theta^k(b|s)\bigr)\bigl(\mathbf 1\{u=b\}-\pi_\theta^k(b|s)\bigr),
\end{equation}
we obtain
\begin{equation}
\label{eq:dpi-final}
\Delta\pi(c|s) \;=\; \eta\,A(s,u)\,\pi_\theta^k(c|s)\,B(c,u;s) \;+\; O(\eta^2),
\end{equation}
which is the claim of Proposition~\ref{prop:per-token}. \hfill$\square$

\section{Distributional Effects of a Sampled-Token OPD Update}
\label{app:softmax-details}

Section~\ref{sec:dynamics} characterizes the direction and magnitude of token probability change induced by a single sampled-token OPD update under the idealized softmax policy. Here, we provide a more complete characterization of how the same update redistributes probability mass over the full student distribution.

Recall from Proposition~\ref{prop:per-token} that the first-order
probability change of any token $c\in\mathcal V$ is
\begin{equation}
\label{eq:app-movement-factorization}
\Delta\pi(c\mid s)
=
\eta\,
A(s,u)\,
\pi_\theta^k(c\mid s)\,
B(c,u;s)
+
O(\eta^2).
\end{equation}
The first-order probability change depends on three factors: the sampled-token advantage $A(s,u)$, the student probability $\pi_\theta^k(c\mid s)$, and the coupling term $B(c,u;s)$.
We denote the first-order probability change by
\[
\Delta_1\pi(c\mid s)
:=\eta A(s,u)\pi_\theta^k(c\mid s)B(c,u;s),
\qquad
\Delta\pi(c\mid s)=\Delta_1\pi(c\mid s)+O(\eta^2).
\]
The direction statements below refer to $\Delta_1\pi$; when this term
vanishes, higher-order terms may still produce a nonzero actual change.

For the sampled token $c=u$, the direction and magnitude admit
the simple characterization given in Section~\ref{sec:dynamics}.
We therefore focus below on the non-sampled tokens, for which the
same sampled-token update induces a richer redistribution over the
vocabulary.

\subsection{Non-Sampled Token Redistribution}
\label{app:nonsampled-redistribution}

For a non-sampled token $c\neq u$, the coupling term in
Eq.~\eqref{eq:app-movement-factorization} reduces to
\begin{equation}
\label{eq:app-B-nonsampled}
B(c,u;s)
=
\tau(u;s)
-
\pi_\theta^k(c\mid s),
\end{equation}
where
\begin{equation}
\label{eq:app-tau}
\tau(u;s)
:=
\left\|
\pi_\theta^k(\cdot\mid s)
\right\|_2^2
-
\pi_\theta^k(u\mid s).
\end{equation}
We refer to $\tau(u;s)$ as the \emph{watershed}. It is determined by
the current student distribution and the sampled token: the term
$\pi_\theta^k(u\mid s)$ captures the probability of the sampled
token, while
$\|\pi_\theta^k(\cdot\mid s)\|_2^2$ captures the concentration of the
full distribution.

Substituting Eq.~\eqref{eq:app-B-nonsampled} into
Eq.~\eqref{eq:app-movement-factorization} gives
\begin{equation}
\label{eq:app-nonsampled-update}
\Delta\pi(c\mid s)
=
\eta A(s,u)\,
\pi_\theta^k(c\mid s)
\left[
\tau(u;s)-\pi_\theta^k(c\mid s)
\right]
+
O(\eta^2),
\qquad c\neq u.
\end{equation}
Equation~\eqref{eq:app-nonsampled-update} shows that the probability change of a non-sampled token depends on the sampled-token advantage, its own student probability, and the full student distribution.

\paragraph{Direction.}
Because $\pi_\theta^k(c\mid s)>0$, the first-order direction of a
non-sampled token is determined by the sign of $A(s,u)$ and its
position relative to the watershed:
\begin{equation}
\label{eq:app-watershed-sign}
\operatorname{sign}
\left(
\Delta_1\pi(c\mid s)
\right)
=
\begin{cases}
+\operatorname{sign}\left(A(s,u)\right),
&
\pi_\theta^k(c\mid s)<\tau(u;s),
\\[3pt]
0,
&
\pi_\theta^k(c\mid s)=\tau(u;s),
\\[3pt]
-\operatorname{sign}\left(A(s,u)\right),
&
\pi_\theta^k(c\mid s)>\tau(u;s),
\end{cases}
\end{equation}
to first order in $\eta$.

Thus, the sampled-token advantage provides a common direction signal,
while the current student distribution determines whether each
non-sampled token follows or opposes that signal. A token
lying exactly at the watershed has no first-order probability change.

\paragraph{Magnitude.}
The corresponding first-order magnitude is
\begin{equation}
\label{eq:app-nonsampled-amplitude}
\left|
\Delta\pi(c\mid s)
\right|
=
\eta
\left|A(s,u)\right|
\pi_\theta^k(c\mid s)
\left|
\tau(u;s)-\pi_\theta^k(c\mid s)
\right|
+
O(\eta^2).
\end{equation}
A large sampled-token advantage does not necessarily produce a large probability change for every token. The first-order change also depends on the token's student probability and its distance from the watershed. It is small when the student probability is near zero or close to the watershed.

The watershed therefore controls both aspects of the non-sampled
dynamics: crossing the watershed reverses the first-order direction,
while approaching the watershed reduces the first-order magnitude.

\subsection{Geometry of the Redistribution}
\label{app:redistribution-geometry}

We next characterize the location of the watershed relative to the
current student distribution. This geometry determines how the sampled-token advantage affects the probabilities of non-sampled tokens.

Let
\begin{equation}
c^\star
:=
\arg\max_{b\in\mathcal V}
\pi_\theta^k(b\mid s)
\end{equation}
denote the highest-probability student token.

\begin{theorem}[Mode-direction invariance]
\label{thm:app-mode-direction}
For every sampled token $u\in\mathcal V$,
\begin{equation}
\pi_\theta^k(c^\star\mid s)
>
\tau(u;s).
\end{equation}
Consequently, whenever $c^\star\neq u$,
\begin{equation}
\operatorname{sign}
\left(
\Delta_1\pi(c^\star\mid s)
\right)
=
-
\operatorname{sign}
\left(
A(s,u)
\right)
\end{equation}
to first order in $\eta$.
\end{theorem}

\begin{proof}
Since
\[
\pi_\theta^k(b\mid s)
\le
\pi_\theta^k(c^\star\mid s)
\qquad
\text{for every } b\in\mathcal V,
\]
we have
\begin{align}
\left\|
\pi_\theta^k(\cdot\mid s)
\right\|_2^2
&=
\sum_{b\in\mathcal V}
\pi_\theta^k(b\mid s)^2
\le
\pi_\theta^k(c^\star\mid s)
\sum_{b\in\mathcal V}
\pi_\theta^k(b\mid s)
=
\pi_\theta^k(c^\star\mid s).
\end{align}
Therefore,
\begin{align}
\tau(u;s)
&=
\left\|
\pi_\theta^k(\cdot\mid s)
\right\|_2^2
-
\pi_\theta^k(u\mid s)
\le
\pi_\theta^k(c^\star\mid s)
-
\pi_\theta^k(u\mid s)
<
\pi_\theta^k(c^\star\mid s),
\end{align}
where the strict inequality follows from
$\pi_\theta^k(u\mid s)>0$ under the softmax parameterization.
\end{proof}

Theorem~\ref{thm:app-mode-direction} shows that whenever the student
mode is not itself sampled, it always lies above the watershed and
therefore moves against the sampled-token advantage.

\paragraph{Sampling the student mode.}

The redistribution simplifies further when the sampled token is itself
the student mode.

\begin{corollary}[Mode sampling]
\label{cor:app-mode-sampling}
If $u=c^\star$, then
\begin{equation}
\tau(u;s)\le0.
\end{equation}
Consequently, every non-sampled token satisfies
\begin{equation}
\operatorname{sign}
\left(
\Delta_1\pi(c\mid s)
\right)
=
-
\operatorname{sign}
\left(
A(s,u)
\right),
\qquad c\neq u,
\end{equation}
to first order in $\eta$.
\end{corollary}

\begin{proof}
If $u=c^\star$, then
\[
\pi_\theta^k(b\mid s)
\le
\pi_\theta^k(u\mid s)
\qquad
\text{for every } b\in\mathcal V.
\]
Hence,
\begin{align}
\left\|
\pi_\theta^k(\cdot\mid s)
\right\|_2^2
&=
\sum_{b\in\mathcal V}
\pi_\theta^k(b\mid s)^2
\le
\pi_\theta^k(u\mid s)
\sum_{b\in\mathcal V}
\pi_\theta^k(b\mid s)
=
\pi_\theta^k(u\mid s),
\end{align}
and therefore
\begin{equation}
\tau(u;s)
=
\left\|
\pi_\theta^k(\cdot\mid s)
\right\|_2^2
-
\pi_\theta^k(u\mid s)
\le0.
\end{equation}
Since every non-sampled softmax probability is strictly positive,
\[
\pi_\theta^k(c\mid s)>\tau(u;s)
\qquad
\text{for every } c\neq u,
\]
and the result follows from
Eq.~\eqref{eq:app-watershed-sign}.
\end{proof}

Thus, mode sampling induces a particularly simple distribution-wide
redistribution: the sampled mode moves with its advantage, while all
remaining tokens move in the opposite direction.

\paragraph{Redistribution regimes.}

More generally, the sign of the watershed distinguishes two
qualitatively different redistribution regimes.

\emph{Non-positive watershed: $\tau(u;s)\le0$.}
Since every softmax probability is strictly positive,
\begin{equation}
\pi_\theta^k(c\mid s)>\tau(u;s),
\qquad
c\neq u.
\end{equation}
Hence all non-sampled tokens move against the sampled-token advantage:
\begin{equation}
\operatorname{sign}
\left(
\Delta_1\pi(c\mid s)
\right)
=
-
\operatorname{sign}
\left(
A(s,u)
\right),
\qquad
c\neq u,
\end{equation}
to first order. Mode sampling always belongs to this regime, although
a non-positive watershed can also occur when an off-mode token is
sampled.

\emph{Positive watershed: $\tau(u;s)>0$.}
Non-sampled tokens may lie on either side of the watershed.
Tokens satisfying
\begin{equation}
\pi_\theta^k(c\mid s)<\tau(u;s)
\end{equation}
move with the sampled-token advantage, whereas tokens satisfying
\begin{equation}
\pi_\theta^k(c\mid s)>\tau(u;s)
\end{equation}
move against it. Tokens satisfying
\begin{equation}
\pi_\theta^k(c\mid s)=\tau(u;s)
\end{equation}
have zero first-order probability change.

Thus, when the watershed is positive, non-sampled token probabilities can change in different directions. However, a given distribution need not have non-sampled tokens on both sides of the watershed.

\subsection{Probability Changes Relative to the Teacher}
\label{app:teacher-direction}

The preceding analysis characterizes whether a non-sampled token moves
with or against the \emph{sampled-token advantage}. This is distinct
from asking whether that token moves toward its own probability under
the teacher.

For each token $c$, define the teacher-preferred direction as
\begin{equation}
\label{eq:app-teacher-direction}
D_T(c;s)
:=
\operatorname{sign}
\left(
\pi_T(c\mid s)
-
\pi_\theta^k(c\mid s)
\right).
\end{equation}
For a token whose student and teacher probabilities differ, the
first-order change points toward the teacher when
\begin{equation}
\label{eq:app-teacher-aligned}
\operatorname{sign}
\left(
\Delta_1\pi(c\mid s)
\right)
D_T(c;s)>0,
\end{equation}
and away from the teacher when this product is negative. If
$\Delta_1\pi(c\mid s)=0$, there is no first-order movement. If
$D_T(c;s)=0$, the probabilities already match, and any nonzero
first-order change moves the student away from that match.

For the sampled token $u$, these two notions coincide. Since the
logarithm is monotone,
\begin{equation}
\operatorname{sign}
\left(
A(s,u)
\right)
=
\operatorname{sign}
\left(
\pi_T(u\mid s)
-
\pi_\theta^k(u\mid s)
\right).
\end{equation}
Together with the sampled-token direction established in
Section~\ref{sec:dynamics}, this implies that the sampled token
moves toward its corresponding teacher probability to first order
when the probabilities differ, and remains unchanged when they match.

For a non-sampled token $c\neq u$, however, the first-order update
direction is
\begin{equation}
\label{eq:app-update-vs-teacher}
\operatorname{sign}
\left(
\Delta_1\pi(c\mid s)
\right)
=
\operatorname{sign}
\left(
A(s,u)
\right)
\operatorname{sign}
\left(
\tau(u;s)-\pi_\theta^k(c\mid s)
\right),
\end{equation}
which need not coincide with
$D_T(c;s)$. Equivalently, a non-sampled token is moved toward the
teacher when
\begin{equation}
\label{eq:app-alignment-condition}
\operatorname{sign}
\left(
A(s,u)
\right)
\operatorname{sign}
\left(
\tau(u;s)-\pi_\theta^k(c\mid s)
\right)
D_T(c;s)
>0,
\end{equation}
and away from the teacher when the product is negative.

Equation~\eqref{eq:app-alignment-condition} also gives a geometric
interpretation in the student--teacher probability plane. The
diagonal
\begin{equation}
\pi_T(c\mid s)
=
\pi_\theta^k(c\mid s)
\end{equation}
separates the two teacher-preferred directions, while the vertical
boundary
\begin{equation}
\pi_\theta^k(c\mid s)
=
\tau(u;s)
\end{equation}
separates the two directions induced by the sampled-token update.
Together, these boundaries partition the plane into regions where the
induced probability change is either aligned with or opposed to the teacher.

Consequently, a single sampled-token OPD step need not move every
token toward its corresponding teacher probability. The sampled
token receives a teacher-directed correction, whereas the
remaining tokens respond indirectly through the redistribution
induced by the sampled-token advantage and the current student
distribution.

\subsection{Empirical Validation of Proposition~\ref{prop:per-token}}

In this subsection, we validate the correctness of Proposition~\ref{prop:per-token} using a categorical policy with $|\mathcal V|=100$. We randomly initialize student and teacher logits and obtain their categorical distributions through softmax. For each trial, we sample $u\sim\pi_\theta(\cdot\mid s)$ and apply one SGD update.

For each update, we compute the predicted changes in log probability and probability from Proposition~\ref{prop:per-token}, using the pre-update distributions. We measure the actual changes by subtracting the pre-update values from the post-update values, as defined in~\eqref{eq:delta_pi}, and compare the predictions with these measurements across all tokens and random initializations. The close agreement with the identity line validates the per-token dynamics in Proposition~\ref{prop:per-token}.

\begin{figure}
    \centering
    \includegraphics[width=1.0\linewidth]{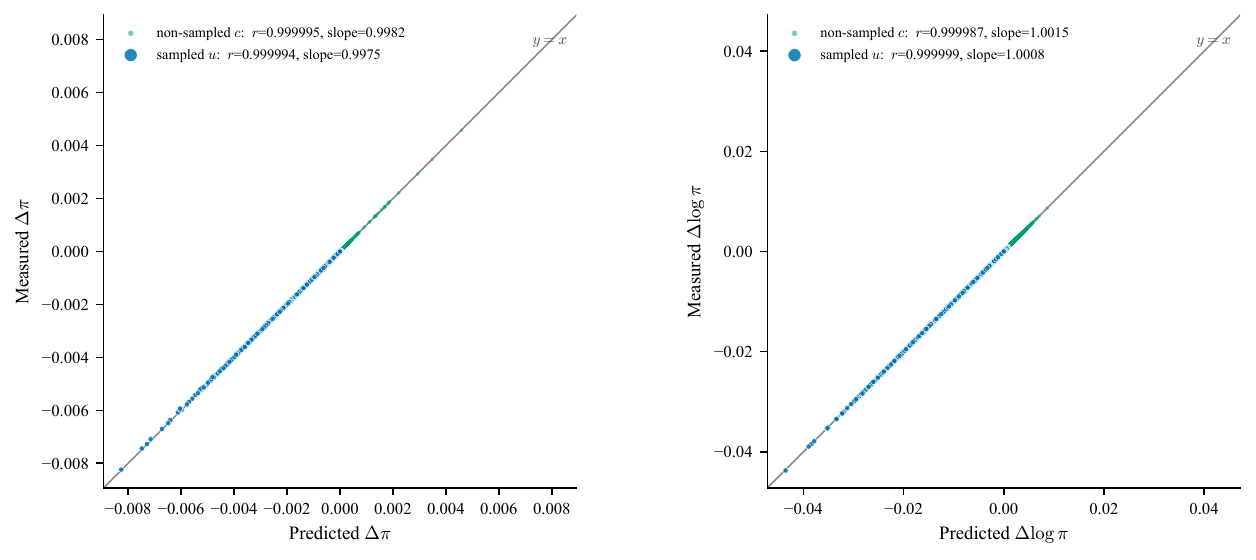}
    \caption{
\textbf{Validation of Proposition~\ref{prop:per-token} under randomly initialized categorical policies.}
Predicted and measured changes are shown for
$\Delta\log\pi$ and $\Delta\pi$.
Each point represents one vocabulary token after a single
sampled-token update; the dashed line denotes $y=x$.
}
    \label{fig:placeholder}
\end{figure}

\subsection{Controlled Token-Level Interventions}
\label{app:controlled-token-interventions}

The preceding analysis separates the direction induced by the
sampled-token update from the direction preferred by the teacher.
We next isolate the sources of this behavior through two controlled
interventions. First, we vary the sampled token while holding both the
student and teacher distributions fixed. We then vary the teacher
distribution while holding the student distribution and sampled token
fixed.

\paragraph{Varying the sampled token under fixed student and teacher
distributions.}

We first hold
$\pi_\theta^k(\cdot\mid s)$ and $\pi_T(\cdot\mid s)$ fixed and vary
only the sampled token $u$. This isolates the effect of the sampled
token on the one-step redistribution.

Changing $u$ modifies the update through two quantities. First, it
changes the sampled-token advantage
\begin{equation}
A(s,u)
=
\log\pi_T(u\mid s)
-
\log\pi_\theta^k(u\mid s).
\end{equation}
Second, it changes the watershed
\begin{equation}
\tau(u;s)
=
\left\|
\pi_\theta^k(\cdot\mid s)
\right\|_2^2
-
\pi_\theta^k(u\mid s).
\end{equation}
Thus, even with the same teacher and student distributions, changing the sampled token can change both the advantage $A(s,u)$ and the watershed $\tau(u;s)$.

\begin{figure*}[t]
    \centering
    \includegraphics[width=\textwidth]{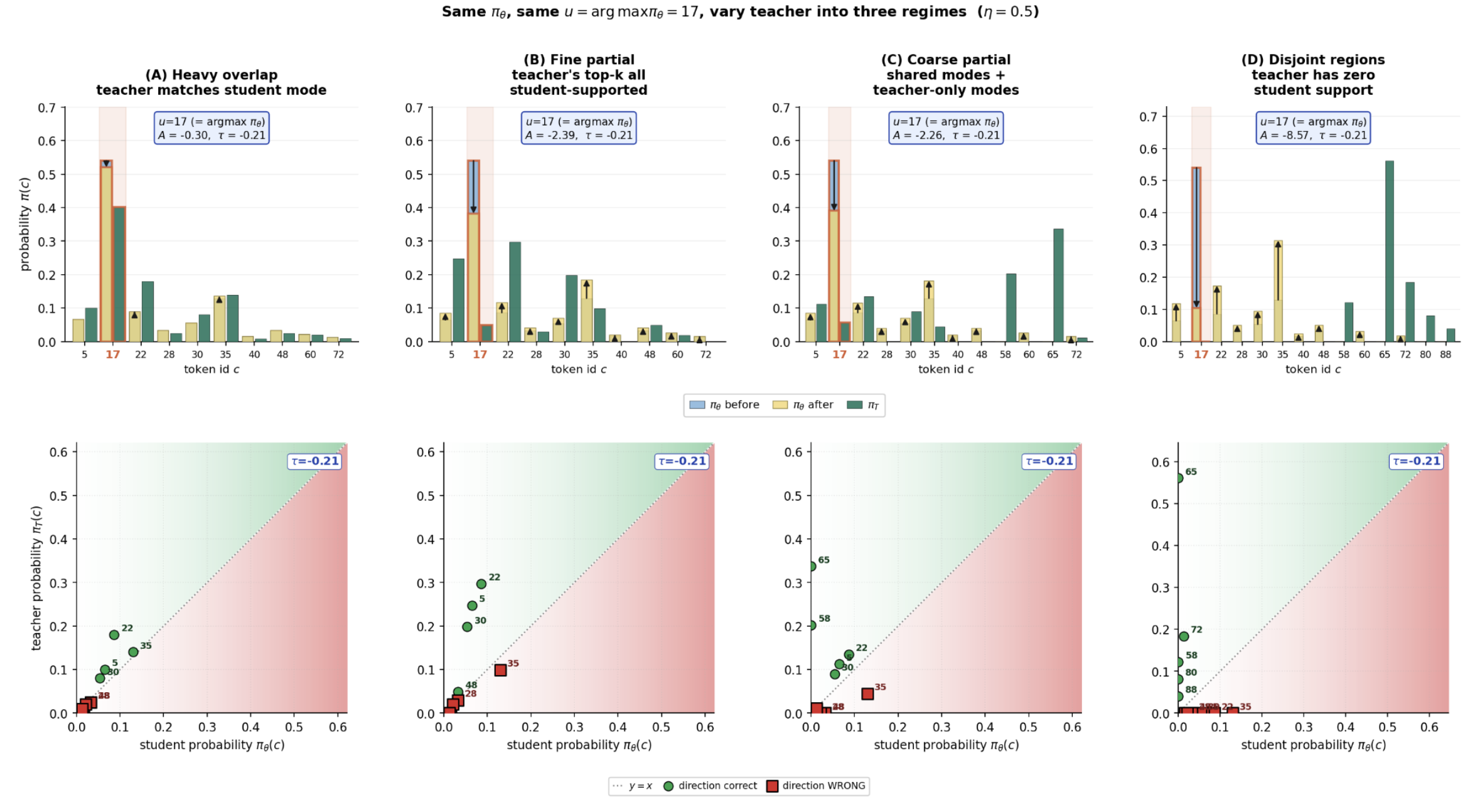}
    \caption{
    \textbf{Changing the sampled token changes the one-step
    redistribution under fixed student and teacher distributions.}
    Across all columns,
    $\pi_\theta^k(\cdot\mid s)$ and $\pi_T(\cdot\mid s)$ are fixed;
    only the sampled token $u$ changes.
    \emph{Top:} the student distribution before and after one OPD
    update together with the teacher distribution. The sampled token is highlighted.
    \emph{Bottom:} each token is plotted according to its student and
    teacher probabilities. The diagonal denotes equal student and
    teacher probability, while the vertical line marks the watershed
    $\tau(u;s)$. Green and red regions indicate probability changes toward and
    away from the teacher, respectively.
    Changing the sampled token modifies both $A(s,u)$ and
    $\tau(u;s)$, and can therefore change the teacher alignment of the
    same token even though the teacher and student
    distributions are unchanged.
    }
    \label{fig:app-vary-sampled-token}
\end{figure*}

Figure~\ref{fig:app-vary-sampled-token} shows that the identity of the
sampled token can qualitatively change the resulting redistribution.
For example, the same token can move away from its
teacher probability under one sampled token but toward the teacher
under another. In the illustrated cases, token $35$ exhibits such a
change despite both the teacher and student distributions remaining
fixed.

The difference arises entirely from the sampled-token-dependent terms
in Eq.~\eqref{eq:app-nonsampled-update}: changing $u$ changes both the
sign and magnitude of $A(s,u)$ and the location of the watershed
$\tau(u;s)$. Thus, the teacher--student pair alone does not determine
the one-step probability change of the full distribution; the sampled token is
an essential part of the dynamics.

\paragraph{Varying the teacher while fixing the student distribution and sampled token.}

We next fix the student distribution and sampled token $u$, while
varying only the teacher distribution. Since
$\pi_\theta^k(\cdot\mid s)$ and $u$ are unchanged, the student-side
watershed
\begin{equation}
\tau(u;s)
=
\left\|
\pi_\theta^k(\cdot\mid s)
\right\|_2^2
-
\pi_\theta^k(u\mid s)
\end{equation}
is identical across all cases. The student probability $\pi_\theta^k(c\mid s)$ of each token $c$ is also unchanged.

Changing the teacher nevertheless affects the dynamics in two ways.
It changes the sampled-token advantage $A(s,u)$, which controls the
common sign and scale of the induced redistribution, and it changes
the teacher-preferred direction
$D_T(c;s)$ of each token. This intervention therefore
holds the student-side geometry fixed while varying the structure of
the teacher--student mismatch.

We consider teacher distributions with progressively less probability-mass overlap with the student.

\begin{figure*}[t]
    \centering
    \includegraphics[width=\textwidth]{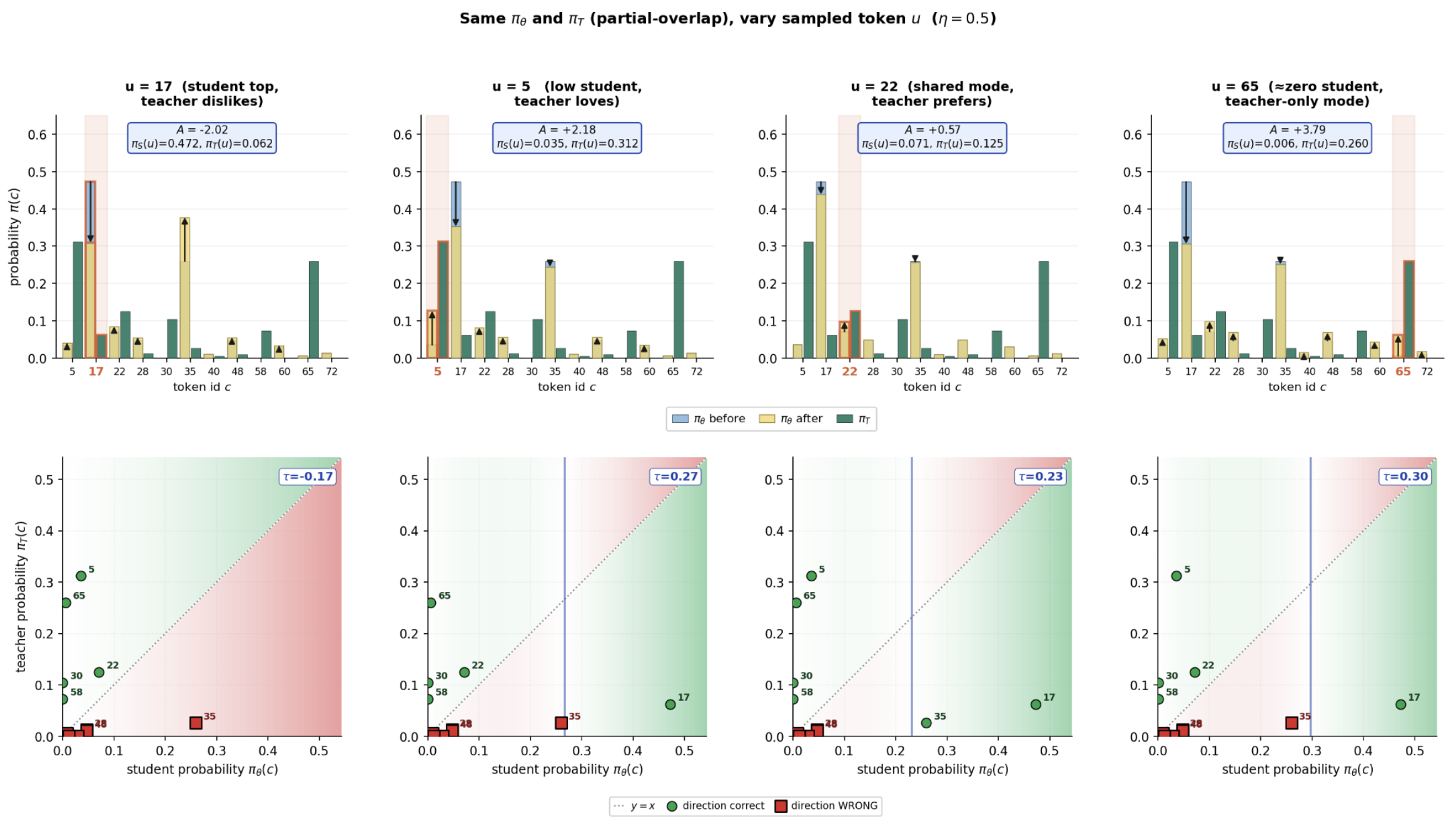}
    \caption{
    \textbf{Teacher alignment changes with the mismatch structure under
    fixed student-side geometry.}
    Across all columns, the student distribution and sampled token $u$
    are fixed, while the teacher distribution has progressively less probability-mass overlap with the student.
    Consequently, the watershed $\tau(u;s)$ and the
    student token probabilities remain unchanged, whereas the
    sampled-token advantage $A(s,u)$ and the teacher-preferred
    directions vary.
    \emph{Top:} the student distribution before and after one OPD
    update together with each teacher distribution.
    \emph{Bottom:} token-level student and teacher probabilities, with
    green and red regions indicating probability changes toward and away from the
    teacher, respectively.
    The same student-side probability geometry can therefore produce
    substantially different patterns of teacher alignment as the
    mismatch structure changes.
    }
    \label{fig:app-vary-teacher}
\end{figure*}

Figure~\ref{fig:app-vary-teacher} illustrates several qualitative
behaviors. When teacher and student probability mass substantially
overlap, much of the induced redistribution can align with the
teacher. Under partial overlap, however, individual non-sampled
tokens can move in conflict with their teacher-preferred
directions; token $35$ provides one such example.

As the teacher and student distributions overlap less, tokens favored by the teacher but assigned very low student probabilities may move toward their teacher probabilities only slightly. Meanwhile, tokens with higher student probabilities may move farther away from their teacher probabilities.

This asymmetry follows directly from the magnitude of the probability change for a non-sampled token
\begin{equation}
\left|
\Delta\pi(c\mid s)
\right|
=
\eta
\left|A(s,u)\right|
\pi_\theta^k(c\mid s)
\left|
\tau(u;s)-\pi_\theta^k(c\mid s)
\right|
+
O(\eta^2).
\end{equation}
Even when the induced direction is teacher-aligned, a token with
near-zero student probability receives little probability change because of the multiplicative factor
$\pi_\theta^k(c\mid s)$. Conversely, a higher-probability token
can undergo a substantially larger probability change even when that probability change
is opposed to the teacher.

These two experiments separate the effects of the sampled token and the teacher distribution. Changing the sampled token changes both $A(s,u)$ and $\tau(u;s)$. Changing the teacher while fixing the student distribution and sampled token leaves $\tau(u;s)$ unchanged, but changes $A(s,u)$ and whether each token's probability change is toward or away from the teacher.

These examples make explicit the factorization in
Eq.~\eqref{eq:app-movement-factorization}: the probability change of
a token is jointly determined by the sampled-token
advantage, its own current probability, and the geometry of the student
distribution. Whether that probability change is toward or away from the teacher
is a separate property determined by how the induced direction aligns
with the teacher--student mismatch at that token.

\section{Relation to Full-Vocabulary OPD}
\label{app:full-vocab}

Taking the expectation of the update in Eq.~\eqref{eq:opd-update} over the sampled token $u\sim\pi_\theta(\cdot|s)$ gives
\begin{equation}
\mathbb E_u\bigl[\Delta\theta_b\bigr] \;=\; \eta\,\pi_\theta(b|s)\bigl(A(s,b)-\bar A(s)\bigr),
\qquad \bar A(s) \;:=\; \mathbb E_{u\sim\pi_\theta(\cdot|s)}\bigl[A(s,u)\bigr],
\end{equation}
which is the gradient step of the full-vocabulary reverse KL at state $s$. Full-vocabulary OPD therefore removes the noise from sampling a single token, at the cost of storing the full-vocabulary logits over the whole sequence. The update of each token is still scaled by its student probability, so teacher-preferred tokens that the student rarely produces still receive little correction.

\begin{figure}[h]
    \centering
    \includegraphics[width=1.0\linewidth]{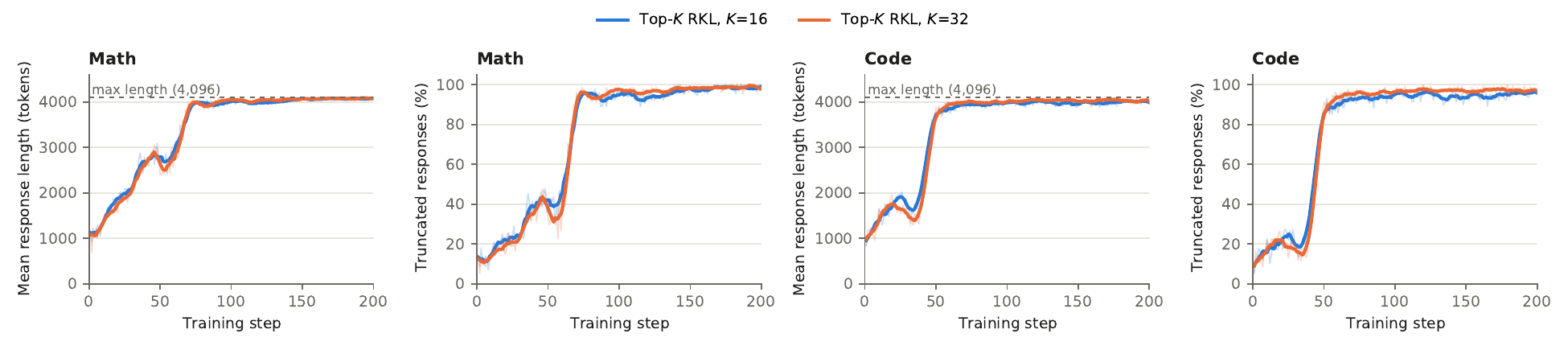}
    \caption{Training response length and fraction of truncated responses for top-$K$ reverse-KL OPD (Qwen3-1.7B-Base ← Qwen3-8B) with $K{=}16$ and $K{=}32$.}
    \label{fig:full_topk_kl}
\end{figure}

To test this, we run OPD with a top-$K$ approximation of the full-vocabulary reverse KL (calculate reverse KL over top-$K$ set, $K=16$) on \texttt{Qwen3-1.7B-Base} with the \texttt{Qwen3-8B} teacher. As shown in Figure~\ref{fig:full_topk_kl}, the same failure appears. On math resoning task, the mean response length grows from 1.2k to 4.1k tokens over 200 steps, and the fraction of rollouts hitting the 4{,}096-token cap rises from 14\% to 100\%. On code, the length grows from 0.8k to 4.0k tokens, and the completion rate rises from 95\% to 5\%. The outputs also become repetitive as shown in Figure~\ref{example1}: the validation loop rate peaks at 81\% on math and 52\% on code, and ends at 42\% and 30\%.

\begin{tcolorbox}[
    colback=blue!5, colframe=blue!50, colbacktitle=blue!50,
    coltitle=white, fonttitle=\bfseries,
    title={Example of a degeneration under topk approximated reverse KL},
    boxrule=0.6pt, arc=2pt, breakable
]
\begin{Verbatim}[breaklines, breakanywhere, fontsize=\scriptsize]
"UIAlertAction ק �� �� ❄ ✅
### Step 1: Understanding the Problem
We are given the set of complex numbers $z$ satisfying the equation:
$$ |1 + z + z^2| = 4 $$
[... Steps 2-7 omitted ...]
### Step 8: Final Answer
$$ m + n = 113 + 2 = \boxed{115} $$
--- ### Final Answer: $$ \boxed{115} $$ �� ### Final Answer: $\boxed{115}$
�� �� ♻ �� �� �� ⚙ �� �� �� �� �� �� �� ��   [repeated until the length limit]"
\end{Verbatim}
\label{example1}
\end{tcolorbox}

\section{Diagnostic Experiments: Setup and Additional Results}

\subsection{Experimental Setup}
\label{app:diagnostic-setup}

We use several teacher--student pairs to examine how OPD
training dynamics vary across student and teacher models. Table~\ref{tab:diagnostic-models} defines the abbreviations used throughout the diagnostic figures. We construct the held-out prefixes from responses generated by the teacher model on the AIME dataset.

\paragraph{Overlap and overlap mass.}
At a state $s$, let $\mathcal S_K^\theta(s)$ denote the student's top-$K$ token set and let $\mathcal S_K(s)$ denote the teacher's top-$K$ token set, as defined in the main text. Their overlap is the intersection
\[
\mathcal I_K(s)=\mathcal S_K^\theta(s)\cap\mathcal S_K(s).
\]
The student and teacher probability masses on this intersection are
\[
M_{\mathrm{overlap}}^\theta(s)
=\sum_{c\in\mathcal I_K(s)}\pi_\theta(c\mid s),
\qquad
M_{\mathrm{overlap}}^T(s)
=\sum_{c\in\mathcal I_K(s)}\pi_T(c\mid s).
\]
We use $K=16$ for these overlap diagnostics. Both masses use the original probabilities without renormalization. Unlike student coverage of the teacher's top-$K$ set, overlap mass sums only over tokens present in both top-$K$ sets.

\begin{table}[t]
\centering
\small
\begin{tabular}{ll}
\toprule
Abbreviation & Model or checkpoint \\
\midrule
Qwen3-1.7B-Base & \texttt{Qwen3-1.7B-Base} \\
Qwen3-1.7B-SFT  & SFT of Qwen3-1.7B-Base on Qwen3-4B \\
R1-Distill-1.5B & \texttt{DeepSeek-R1-Distill-Qwen-1.5B} \\
JustRL-1.5B     & \texttt{hbx/JustRL-DeepSeek-1.5B} \\
Qwen3-4B       & \texttt{Qwen3-4B} \\
Qwen3-8B       & \texttt{Qwen3-8B} \\
\bottomrule
\end{tabular}
\caption{Model abbreviations used in the diagnostic experiments.}
\label{tab:diagnostic-models}
\end{table}

\subsection{Additional Results on Local Distributional Dynamics}
\label{app:token-movement}

We further examine the relationship between student probability and realized token probability change, measured at the same prefixes before and after OPD training.

\begin{figure}[h]
    \centering
    \begin{subfigure}[t]{0.48\linewidth}
        \centering
        \includegraphics[width=\linewidth]{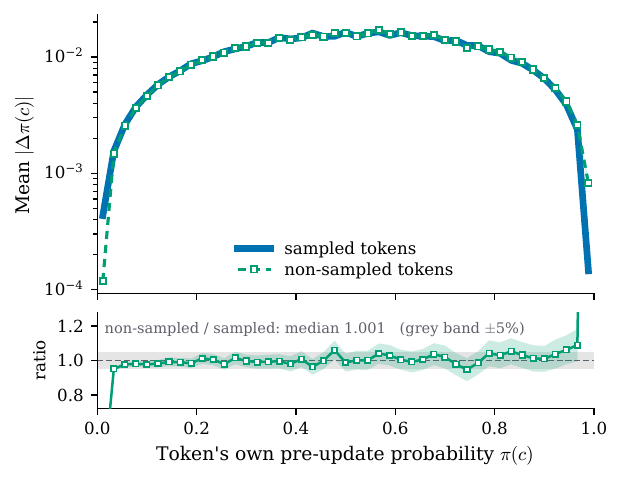}
        \caption{Sampled vs.\ non-sampled tokens.}
    \end{subfigure}
    \hfill
    \begin{subfigure}[t]{0.48\linewidth}
        \centering
        \includegraphics[width=\linewidth]{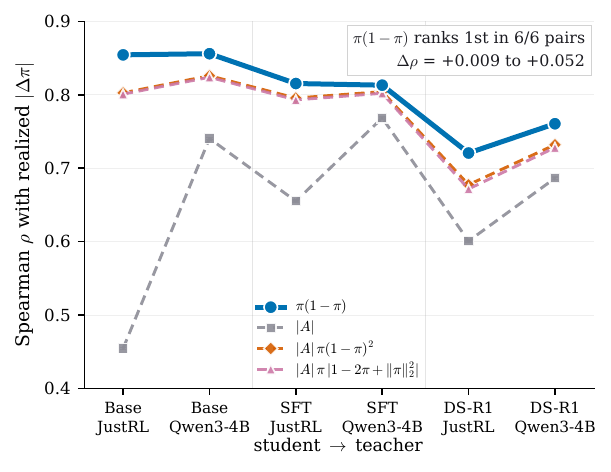}
        \caption{Predictor comparison.}
    \end{subfigure}
    \caption{
    \textbf{Additional comparisons of token probability change under OPD.}
    \textbf{(a)} Probability change for sampled and
    non-sampled tokens, grouped by pre-update student probability.
    \textbf{(b)} Predictive performance of probability-based and
    advantage-based quantities for realized $|\Delta\pi|$.  
    }
    \label{fig:additional-token-movement}
\end{figure}

Figure~\ref{fig:additional-token-movement}(a) shows that sampled and non-sampled tokens exhibit similar magnitudes of change $|\Delta\pi|$ at comparable student probabilities. Non-sampled tokens are selected from the union of the student and teacher top-$K$ sets. Figure~\ref{fig:additional-token-movement}(b) shows that probability-based measures predict $|\Delta\pi|$ more accurately than the advantage-based measures considered. Therefore, these results show that realized probability change depends on the student's current distribution, beyond the sampled-token advantage alone.

\subsection{Validation of the Cross-Position Decomposition}
\label{app:interaction-validation}

\begin{figure}[h]
    \centering
    \includegraphics[width=0.48\linewidth]{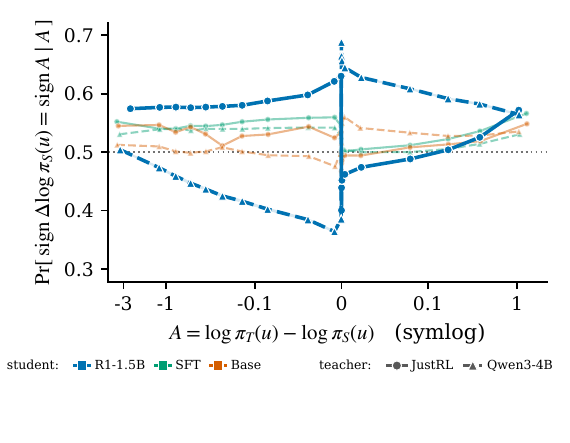}
    \hfill
    \includegraphics[width=0.48\linewidth]{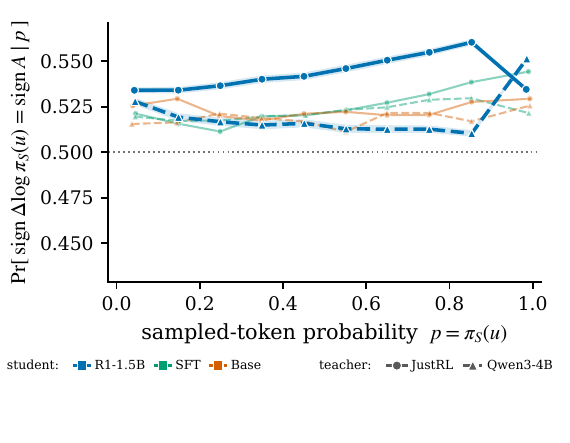}
    \caption{
        Sign agreement between the student's probability
        change and the advantage, conditioned on
        (a) advantage magnitude and
        (b) the student's pre-update probability. Sign agreement remains near $50\%$ across both factors.
    }
    \label{fig:appendix:factor_sign}
\end{figure}

We further examine whether sign agreement varies with
the magnitude of the advantage or the student's
pre-update probability. As shown in
Figure~\ref{fig:appendix:factor_sign}, sign agreement
remains near $50\%$ across both factors, with no
clear dependence on either.

\begin{figure}[h]
\centering
\includegraphics[width=\linewidth]{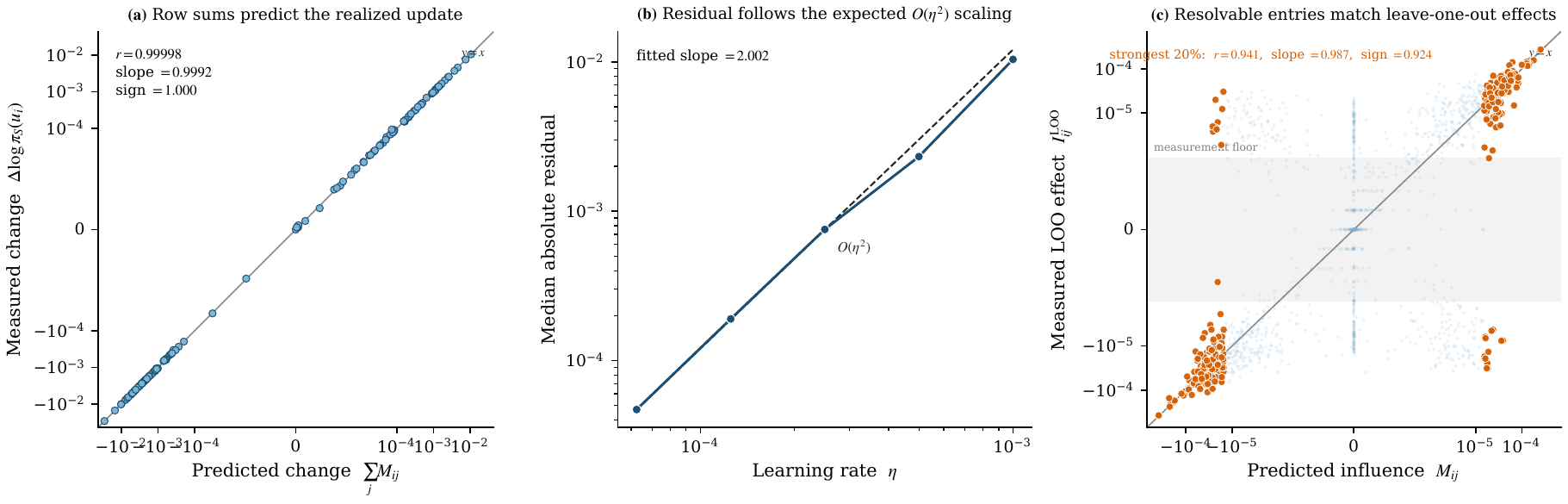}
\caption{
\textbf{Validation of the cross-position influence decomposition.}
\textbf{(a)} The row sum $\sum_j M_{ij}$ closely matches the measured one-step change $\Delta\log\pi_S(u_i)$.
\textbf{(b)} The approximation residual decreases as $O(\eta^2)$ with the learning rate, consistent with a first-order expansion.
\textbf{(c)} Individual $M_{ij}$ values agree with leave-one-out marginal effects when the interaction magnitude is above the numerical measurement floor; the strongest $20\%$ by $|M_{ij}|$ are highlighted.
}
\label{fig:interaction_validation}
\end{figure}

\paragraph{Predicting the log-probability change from row sums.}
Under the first-order approximation, the realized change at receiver position $i$ is
$$
\Delta\log\pi_S(u_i) = \sum_j M_{ij} +O(\eta^2).
$$
As shown in Fig.~\ref{fig:interaction_validation}(a), the row sum $\widehat{\Delta}_i=\sum_j M_{ij}$ closely matches the measured one-step change in sampled-token log probability after an actual SGD update. This confirms that the interaction matrix captures the realized joint update to first order.

\paragraph{Residual scaling with step size.}
We repeat the measurement at different learning rates to check whether the prediction error follows the expected $O(\eta^2)$ scaling. Fig.~\ref{fig:interaction_validation}(b) shows that the residual between $\widehat{\Delta}_i$ and the measured update decreases as $O(\eta^2)$. The observed scaling is therefore consistent with the first-order Taylor approximation underlying the decomposition.

\begin{figure}[h]
    \centering
    \includegraphics[width=1.0\linewidth]{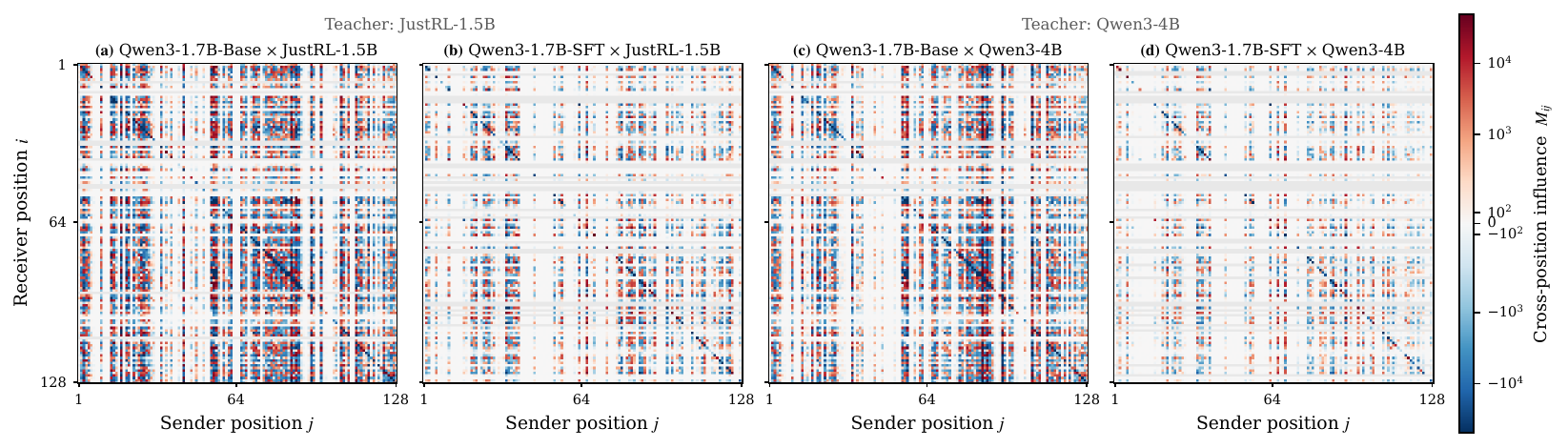}
    \caption{
\textbf{Cross-position interaction structure depends on both the student state and the teacher.}
Each panel shows the exact interaction matrix $M_{ij}$ for one teacher--student pair, with sender position $j$ on the x-axis and receiver position $i$ on the y-axis. Positive and negative influences are shown on a shared symmetric scale.
}
    \label{fig:cross-term}
\end{figure}

\paragraph{Cross-position interactions depend on both the student and the teacher.}
Having shown that cross-position contributions can override a token's own update, we examine how these contributions vary across teacher--student pairs.
Figure~\ref{fig:cross-term} shows substantial off-diagonal
entries in all four settings, with different patterns and magnitudes across Base and SFT students and their teachers. These differences reflect the two factors in
$M_{ij}=\eta A_j\langle h_i,h_j\rangle$:
the student's gradient alignment determines how positions are coupled, while the teacher-dependent advantage weights each sender's contribution.
Thus, the realized update at a position depends on both
its coupling to other positions and the supervision
those positions receive.

\begin{table}[t]
\centering
\scriptsize
\setlength{\tabcolsep}{2pt}
\renewcommand{\arraystretch}{1.08}

\begin{minipage}[t]{0.32\linewidth}
\centering
\vspace{0pt}
\textbf{Step 10}\par\smallskip
\begin{tabular}{@{}rlrr@{}}
\toprule
Rank & Token & $A$ & Weight \\
\midrule
1 & 
                   \texttt{U+0627 U+0643}
  & $-40.38$ & 57.30\% \\
2  & \texttt{paque}          & $-9.00$  & 17.88\% \\
3  & \texttt{tens}           & $-5.14$  & 4.32\% \\
4  & \texttt{<|endoftext|>}   & $-21.16$ & 2.83\% \\
5  & \texttt{a}              & $-1.41$  & 1.96\% \\
6 & \texttt{U+653F U+5E9C} & $-1.99$ & 1.67\% \\
7  & \texttt{written}        & $1.74$   & 1.44\% \\
8  & \texttt{The}            & $-3.04$  & 1.28\% \\
9  & \texttt{at}             & $-1.79$  & 1.16\% \\
10 & \texttt{\$}             & $-1.00$  & 1.08\% \\
\bottomrule
\end{tabular}
\end{minipage}
\hfill
\begin{minipage}[t]{0.32\linewidth}
\centering
\vspace{0pt}
\textbf{Step 50}\par\smallskip
\begin{tabular}{@{}rlrr@{}}
\toprule
Rank & Token & $A$ & Weight \\
\midrule
1  & \texttt{But}            & $-2.81$  & 12.05\% \\
2  & \texttt{\textbackslash(}& $-6.07$  & 8.10\% \\
3  & \texttt{pairwise}       & $-5.06$  & 7.78\% \\
4  & \texttt{,}              & $-3.19$  & 5.20\% \\
5  & \texttt{dots}           & $-8.99$  & 4.95\% \\
6  & \texttt{pair}           & $1.39$   & 4.92\% \\
7  & \texttt{But}            & $-2.91$  & 4.64\% \\
8  & \texttt{positive}       & $-2.18$  & 4.33\% \\
9  & \texttt{circle}         & $-1.44$  & 3.70\% \\
10 & \texttt{7}              & $1.78$   & 3.08\% \\
\bottomrule
\end{tabular}
\end{minipage}
\hfill
\begin{minipage}[t]{0.32\linewidth}
\centering
\vspace{0pt}
\textbf{Step 100}\par\smallskip
\begin{tabular}{@{}rlrr@{}}
\toprule
Rank & Token & $A$ & Weight \\
\midrule
1  & \texttt{<|endoftext|>}   & $-22.53$ & 31.81\% \\
2  & \texttt{Rightarrow}     & $3.76$   & 10.09\% \\
3  & \texttt{that}           & $2.09$   & 6.20\% \\
4  & \texttt{become}         & $-2.61$  & 3.55\% \\
5  & \texttt{terms}          & $-1.35$  & 3.03\% \\
6  & \texttt{value}          & $-1.96$  & 2.71\% \\
7  & \texttt{which}          & $-1.52$  & 2.59\% \\
8  & \texttt{2}              & $-0.89$  & 2.16\% \\
9  & \texttt{let}            & $-0.85$  & 2.07\% \\
10 & \texttt{(}              & $-1.39$  & 2.02\% \\
\bottomrule
\end{tabular}
\end{minipage}

\caption{Top ten tokens by absolute cross-position influence $\sum_{i\neq j}|M_{ij}|$
at three checkpoints.
Ranks and weights are computed separately within each
checkpoint's 200 selected positions, including one
explicitly included termination position.
Repeated token strings denote different positions.
Non-Latin tokens are represented by their Unicode code points.
}
\label{tab:interaction-full}
\end{table}

\paragraph{Leave-one-out validation of individual interactions.}
The row-sum result validates the aggregate decomposition but does not by itself establish that an individual entry $M_{ij}$ can be interpreted as the influence of sender $j$ on receiver $i$. We therefore compare $M_{ij}$ with a leave-one-out intervention in which sender $j$ is removed from the update and the resulting change at receiver $i$ is measured directly.

As shown in Fig.~\ref{fig:interaction_validation}(c), interactions large enough to measure reliably closely match their leave-one-out effects. For the strongest $20\%$ of interactions by $|M_{ij}|$, we obtain Pearson $r=0.941$, slope $=0.987$, and sign agreement $=0.924$. Weaker interactions are too small to measure reliably given the numerical precision of the forward pass. These results support interpreting $M_{ij}$ as the first-order sender-to-receiver influence of position $j$ on position $i$.

\subsection{Interventions on Mismatched Positions}
\label{app:interventions}

This section gives the details of the interventions in Section~\ref{sec:sampling-dynamics}. All runs use the setting of Section~\ref{sec:sampling-dynamics}: a \texttt{Qwen3-1.7B-Base} student, a \texttt{Qwen3-8B} teacher, the DAPO dataset, and 200 training steps with sampled-token OPD.

\paragraph{Controls on the termination token.}
In the original setting (A), the student ends a response with \texttt{<|endoftext|>}, while the teacher uses \texttt{<|im\_end|>}. We compare it with two controls. In (B), we align the termination tokens by mapping the student's termination token to the teacher's when computing the advantage. In (C), we keep the mismatch but mask the gradient at positions where the sampled token is a termination token. Table~\ref{tab:eos-controls} reports the results. Both controls remove the collapse: the fraction of rollouts that hit the length cap drops from 0.95 to 0.25, and the validation response length drops from 7{,}234 to below 4{,}000 tokens.

\begin{table}[h]
\centering
\small
\begin{tabular}{lccc}
\toprule
& A: original & B: aligned & C: gradient masked \\
\midrule
Rollouts hitting the Max length & 0.95 & 0.25 & 0.25 \\
Validation response length (tokens) & 7{,}234 & 3{,}939 & 3{,}761 \\
Top-16 overlap, teacher mass & 0.988 & 0.989 & $\approx$0.99 \\
\bottomrule
\end{tabular}
\caption{Controls on the termination token.}
\label{tab:eos-controls}
\end{table}

\paragraph{Injected mismatch on ordinary tokens.}
To test whether a constant mismatch, where leading to large negative advantage alone, causes failure, we start from the aligned setting (B) and inject a persistent mismatch on an ordinary token: whenever the student samples this token, we set the teacher's log-probability of it to -10, which gives a large negative advantage. We run this separately for five tokens: \texttt{\#\#\#}, \texttt{---\textbackslash n\textbackslash n}, \texttt{\textbackslash n\textbackslash n}, \texttt{The}, and \texttt{So}. In every run, the student stops producing the penalized token. For example, \texttt{\textbackslash n\textbackslash n} falls from 1.8\% to 0.02\% of generated tokens. Accuracy is slightly lower, with a mean Avg@8 of 16.9--18.3 over AIME24--26 and AMC23, compared with 18.5 without injection.

These results clarify which mismatched tokens matter. When the mismatch is on an ordinary token, the student simply stops producing it and moves to other tokens that the teacher also accepts, so the mismatch is absorbed. The mismatch becomes harmful only when the teacher puts almost all of its probability on a token that the student cannot produce and that the response cannot do without, such as the termination token or format token.

\section{Experimental Settings}
\label{app:experimental-details}

\paragraph{Models and Training Datasets.} We conduct experiments
using three Qwen3~\citep{yang2025qwen3} student models
of different sizes, Qwen3-1.7B-Base, Qwen3-4B-Base,
and Qwen3-8B-Base. We use Qwen3-8B as the teacher
model in both settings:
non-thinking mode for OPD directly from Base initialization,
and thinking mode for OPD from SFT initialization. We use the DAPO dataset~\citep{yu2026dapo} for OPD training
across all student sizes. For SFT, we extract 400K prompt--response pairs from
\texttt{open-thoughts/OpenThoughts3-1.2M}, comprising 50K distinct prompts with eight responses each.

\paragraph{Evaluation Benchmarks and Metrics.} We evaluate on AIME24-26~\citep{maa_aime} and AMC23~\citep{maa2023amc}.
We generate eight responses per question using different validation settings for the Base and SFT-initialized cases, as specified in Table~\ref{tab:opd_hyperparameters}.
We report Avg@8, the mean accuracy across the eight responses, and Best@8, estimated by bootstrap resampling. For each question, we repeatedly draw eight responses with replacement from its eight generated responses and record whether any resampled response is correct. We average these binary outcomes over bootstrap replicates and then over questions. If $c_i$ of the eight original responses to question $i$ are correct, the conditional expectation of its bootstrap score is $1-(1-c_i/8)^8$. Thus, for $N$ questions, the expected bootstrap metric is
\begin{equation*}
\mathrm{Best@8}=\frac{1}{N}\sum_{i=1}^{N}\left[1-\left(1-\frac{c_i}{8}\right)^8\right].
\end{equation*}
Table~\ref{tab:qwen_math_results} reports the bootstrap estimates as percentages. This metric differs from the fraction of questions with at least one correct response among the original eight generations.

\paragraph{Baselines.} We compare our design against three baselines: (i) OPD~\citep{lu2025onpolicydistillation}, standard on-policy distillation with sampled-token reverse KL; (ii) EOPD~\citep{jin2026entropy}, which uses forward KL when the teacher has high entropy; (iii) AOPD~\citep{jia2026asymmetric}, which uses FKL when the sampled-token advantage is non-positive. For the training-dynamics comparison in Figure~\ref{fig:training_dynamics_4b}, we additionally include All-FKL, which applies the same top-$K$ FKL term as in~\eqref{eq:routed-objective} at every token position, corresponding to $r(s_t)=1$ for all $t$. At fixed $K$, All-FKL requires more teacher-target storage than selective routing when targets are retained only at positions receiving FKL. This storage comparison does not establish a reduction in peak training memory, which also depends on the implementation. \citet{jin2026entropy} discuss the memory overhead of full-vocabulary FKL and the linear growth of top-$K$ target storage with $K$.

\paragraph{Implementation}
We implement on-policy distillation with
verl. We use a training batch size of $128$ and a mini-batch size of $128$ to ensure an on-policy setting. The main hyperparameters are summarized in Table \ref{tab:opd_hyperparameters}. For EOPD~\citep{jin2026entropy}, we use a teacher entropy threshold $\tau=1.0$; for AOPD, we use an advantage threshold of 0 to apply FKL.

\begin{table}[h]
\centering
\small
\setlength{\tabcolsep}{8pt}
\begin{tabular}{lcc}
\toprule
\multirow{2}{*}{\textbf{Hyperparameter}}
& \multicolumn{2}{c}{\textbf{OPD}} \\
\cmidrule(lr){2-3}
& \textbf{Base Model} & \textbf{SFT Model} \\
\midrule
Training steps                 & 200 & 100 \\
Train batch size               & 128 & 128 \\
Micro batch size               & 128 & 128 \\
Rollout $n$                    & 1 & 1 \\
Rollout Correction & [0.5, 2.0] & [0.5, 2.0]\\
Maximum prompt length          & 2048 & 2048 \\
Maximum response length        & 4096 & 16384 \\
Validation max response length & 8192 & 16384 \\
Learning rate                  & $1 \times 10^{-6}$ & $1 \times 10^{-6}$ \\
Temperature                    & 1.0 & 1.0 \\
Top-$p$                        & 1.0 & 1.0 \\
Validation temperature         & 1.0 & 0.6 \\
Validation top-$p$             & 0.8 & 0.95 \\
Validation top-$k$             & -1 & 20 \\
Loss aggregation               & \texttt{token-mean} & \texttt{token-mean} \\
\bottomrule
\end{tabular}
\caption{Default hyperparameters for OPD. For state-dependent KL routing, we use $p=0.9$ for the teacher top-$p$ set, $K=16$ for approximate top-$K$ FKL, and thresholds $\epsilon=6$, $\epsilon_2=0.3$, and $\epsilon_3=0.2$ in~\eqref{eq:fkl-vote}--\eqref{eq:routed-objective}.}
\label{tab:opd_hyperparameters}
\end{table}

For SFT, we fine-tune all model parameters on the extracted 400K
prompt--response pairs using LLaMA-Factory~\citep{zheng2024llamafactory}.
The eight responses per prompt are taken directly from
\texttt{OpenThoughts3-1.2M}. The prompt template is:

\begin{tcolorbox}[
    colback=blue!5,
    colframe=blue!50,
    colbacktitle=blue!50,
    coltitle=white,
    fonttitle=\bfseries,
    title={Prompt template},
    boxrule=0.6pt,
    arc=2pt
]
\{Question\} Please reason step by step, and put your final answer within
\verb|\boxed{}|.
\end{tcolorbox}

We format the responses to match the template used by
Qwen3-8B in thinking mode, ensuring consistent formatting
between SFT and subsequent OPD training. The main SFT hyperparameters are summarized in Table \ref{tab:sft_hyperparameters}.

\begin{table}[h]
\centering
\small
\setlength{\tabcolsep}{10pt}
\begin{tabular}{ll}
\toprule
\textbf{Hyperparameter} & \textbf{Value} \\
\midrule
Student model                  & Base model \\
Training objective             & Full-parameter SFT \\
Template                       & \texttt{qwen3} \\
Training epochs                & 1 \\
Sequence length                & 20,000 \\
Per-device batch size          & 8 \\
Gradient accumulation steps    & 1 \\
Batch Size & 512 \\
Learning rate                  & $8 \times 10^{-5}$ \\
LR scheduler                   & Cosine \\
Warmup ratio                   & 0.1 \\
Precision                      & BF16 \\
\bottomrule
\end{tabular}
\caption{SFT hyperparameters for cold-start initialization of the 1.7B, 4B, and 8B base models using \texttt{OpenThoughts3-1.2M}.}
\label{tab:sft_hyperparameters}
\end{table}

\section{Computational Overhead Analysis}
\label{app:overhead}

\begin{figure}[t]
\centering
\includegraphics[width=\linewidth]{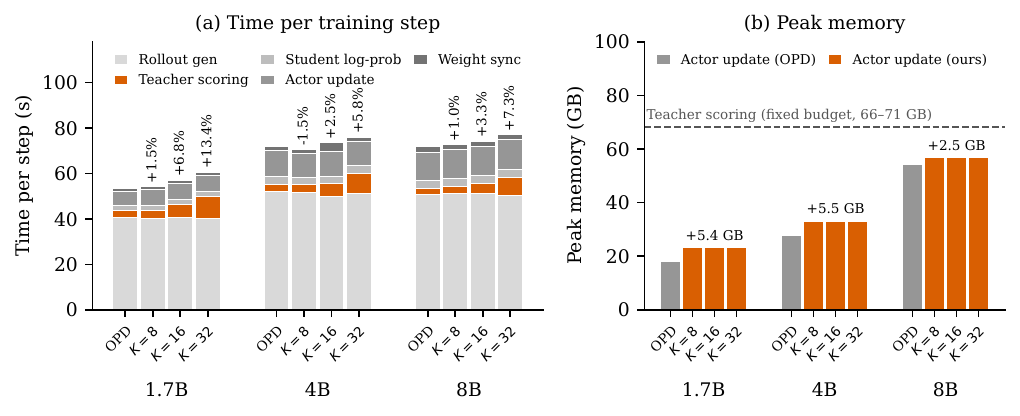}
\caption{Computational overhead of our method compared with sampled-token OPD. (a) Time per training step by stage; labels show the change in total time relative to OPD. (b) Peak memory of the actor update. The dashed line marks the teacher-scoring stage, which runs under a fixed vLLM memory budget and sets the peak memory of a full training step.}
\label{fig:overhead}
\end{figure}

We measure the time and peak memory of each training stage on 4$\times$H100 GPUs with a Qwen3-8B teacher and students of three sizes, averaged over 10 training steps. Teacher scoring is performed with vLLM, with a fixed GPU memory budget for the teacher. Sampled-token OPD retrieves only the teacher's log-probability of the sampled token, while methods that use forward KL, including ours, retrieve the teacher's top-$K$ log-probabilities at each response position. Results are shown in Figure~\ref{fig:overhead}.

\paragraph{Time.} The additional cost comes mostly from teacher scoring and grows roughly linearly with $K$. At $K=16$, teacher scoring increases from 2.6--3.1\,s to 4.5--5.9\,s per step, but it still accounts for only about 6--10\% of the total step time, which is dominated by student rollout generation. The total time per step increases by 2.5--6.8\%.

\paragraph{Memory.} Memory does not grow with $K$ as the teacher-scoring stage runs under a fixed vLLM memory budget. In the actor update, the top-$K$ teacher targets are small ($K$ values per position), and the additional 2.5--5.5\,GB over sampled-token OPD comes from the routing and forward-KL computation, which does not depend on $K$. This stays below the teacher-scoring budget, so the peak memory of a full training step is unchanged.

\section{Ablation Study}
\label{app:ablation}

We examine the routing thresholds, candidate set, and
individual routing rules using Qwen3-1.7B-Base with
the Qwen3-8B teacher. Unless otherwise specified, we use
$\varepsilon_1=6$, $\varepsilon_2=0.3$,
$\varepsilon_3=0.2$, and $p=0.90$.

\paragraph{Routing thresholds.}
Table~\ref{tab:threshold-ablation} shows that increasing
$\varepsilon_1$ from $1.5$ to $6$ improves mean Avg@8
from $16.15\%$ to $18.12\%$, while reducing the final
FKL fraction from $17.6\%$ to $6.7\%$.
Further reducing FKL use does not necessarily improve
performance: increasing $\varepsilon_2$ from $0.3$ to
$0.5$ lowers the FKL fraction to $3.6\%$, but also
reduces mean Avg@8 to $16.90\%$.
Among the tested coverage thresholds, $\varepsilon_3=0.2$
achieves the highest mean Avg@8.
These results show that the choice of routing thresholds
matters beyond the amount of FKL supervision.

\begin{table}[h]
\centering
\small
\setlength{\tabcolsep}{5pt}
\begin{tabular}{llrrrrrr}
\toprule
Parameter & Value & $f_{\mathrm{FKL}}$ (\%)
& AIME24 & AIME25 & AIME26 & AMC23 & Avg. \\
\midrule
\multirow{3}{*}{$\varepsilon_1$}
& 1.5 & 17.6 & 9.17  & 7.50  & 7.92 & 40.00 & 16.15 \\
& 3   & 6.7  & 10.42 & 11.67 & 6.25 & 39.38 & 16.93 \\
& 6   & 6.7 & 12.08 & 9.58 & 5.83 & 45.00 & \textbf{18.12} \\
\midrule
\multirow{4}{*}{$\varepsilon_2$}
& 0.2 & 9.8 & 10.00 & 5.83  & 6.67 & 40.94 & 15.86 \\
& 0.3 & 6.7 & 12.08 & 9.58 & 5.83 & 45.00 & \textbf{18.12} \\
& 0.5 & 3.6 & 9.17  & 8.75  & 7.50 & 42.19 & 16.90 \\
& 0.7 & 1.8 & 10.00 & 5.83  & 7.92 & 39.69 & 15.86 \\
\midrule
\multirow{3}{*}{$\varepsilon_3$}
& 0.1 & 6.9 & 12.50 & 6.67  & 6.67 & 36.56 & 15.60 \\
& 0.2 & 6.7 & 12.08 & 9.58 & 5.83 & 45.00 & \textbf{18.12} \\
& 0.7 & 8.7 & 11.67 & 5.00  & 7.92 & 43.12 & 16.93 \\
\bottomrule
\end{tabular}
\caption{
Parameter sensitivity on Qwen3-1.7B-Base with the
Qwen3-8B teacher. We vary one threshold at a time,
holding the others at $\varepsilon_1=6$,
$\varepsilon_2=0.3$, $\varepsilon_3=0.2$, and $p=0.90$.
Scores are Avg@8 (\%) at each run's best validation
checkpoint over 200 training steps.
$f_{\mathrm{FKL}}$ is the fraction of valid response
positions routed to FKL at the final step.
}
\label{tab:threshold-ablation}
\end{table}

\paragraph{Comparison between $M_{\rm top-p}$ and $M_{\rm top-k}$ on underestimated token}
In Table~\ref{tab:support-ablation}, we compare using the student mass on teacher's adaptive top-$p$ set and top-$k$ set to estimate the underestimated token for $V(\theta;s_t)$.
Thus, using student mass on adaptive teacher top-p
set for routing is more helpful compared with on top-$K$ set , consistent with our local distributional analysis.

\begin{table}[h]
\centering
\small
\setlength{\tabcolsep}{5pt}
\begin{tabular}{lrrrrrr}
\toprule
Candidate set & $f_{\mathrm{FKL}}$ (\%)
& AIME24 & AIME25 & AIME26 & AMC23 & Avg. \\
\midrule
Top-$p$, $p=0.90$
& 6.7 & 12.08 & 9.58 & 5.83 & 45.00 & \textbf{18.12} \\
Top-$p$, $p=0.95$
& 7.1 & 12.92 & 5.83 & 7.92 & 40.94 & 16.90 \\
Fixed top-$K$
& 6.8 & 11.25 & 6.25 & 6.67 & 40.31 & 16.12 \\
\bottomrule
\end{tabular}
\caption{
Comparison between using $M_{\rm top-p}$ and $M_{\rm top-k}$ for mismatch states.
}
\label{tab:support-ablation}
\end{table}

\paragraph{Routing rules.}
Table~\ref{tab:rule-ablation} compares the coverage rule
$M_{\mathrm{top}\text{-}p}$, the underestimation vote
$V$, and their combination.
Combining both rules achieves the highest mean Avg@8
of $18.12\%$, compared with $17.06\%$ for coverage
alone and $16.54\%$ for the vote alone.
The combined rule and vote-only variant have the same
final FKL fraction of $6.7\%$, yet differ by $1.58$
accuracy points.
These results suggest that the two rules provide
complementary criteria for selecting positions for FKL.

\begin{table}[h]
\centering
\small
\setlength{\tabcolsep}{5pt}
\begin{tabular}{lrrrrrr}
\toprule
Active rules & $f_{\mathrm{FKL}}$ (\%)
& AIME24 & AIME25 & AIME26 & AMC23 & Avg. \\
\midrule
$M_{\rm top-p}$ only
& 0.4 & 12.08 & 7.08 & 8.75 & 40.31 & 17.06 \\
$V$ only
& 6.7 & 10.00 & 6.67 & 7.92 & 41.56 & 16.54 \\
$M_{\rm top-p}$ + $V$
& 6.7 & 12.08 & 9.58 & 5.83 & 45.00 & \textbf{18.12} \\
\bottomrule
\end{tabular}
\caption{
Effect of the active routing rules.
}
\label{tab:rule-ablation}
\end{table}
\end{document}